\documentclass[information,article,accept,pdftex,moreauthors]{mdpi} 

\firstpage{1} 
\pubvolume{1}
\issuenum{1}
\articlenumber{0}
\pubyear{2024}
\copyrightyear{2024}
\datereceived{ } 
\daterevised{ } 
\dateaccepted{ } 
\datepublished{ } 
\hreflink{https://doi.org/} 

\Title{Green MLOps to Green GenOps: An Empirical Study of Energy Consumption in Discriminative and Generative AI Operations}

\TitleCitation{Green MLOps to Green GenOps: An Empirical Study of Energy Consumption in Discriminative and Generative AI Operations}

\Author{Adrián Sánchez-Mompó$^{1}$\orcidA{}, Ioannis Mavromatis$^{2,}$*\orcidC{}, Peizheng Li$^{1}$\orcidB{}, Konstantinos Katsaros$^{2}$ and Aftab Khan$^{1,}$*}

\AuthorNames{Adrián Sánchez-Mompó, Ioannis Mavromatis, Peizheng Li, Konstantinos Katsaros, and Aftab Khan}

\AuthorCitation{Sánchez-Mompó, A.; Mavromatis, I.; Li, P.; Katsaros, K.; Khan, A.}

\address{%
$^{1}$ \quad Bristol Research and Innovation Laboratory, Toshiba Europe Ltd., Bristol BS1 4ND,  UK; adrian.mompo@toshiba-bril.com (A.S.-M.); peizheng.li@toshiba-bril.com (P.L.)\\
$^{2}$ \quad {Digital Catapult,}  London NW1 2RA, UK; kostas.katsaros@digicatapult.org.uk}

\corres{Correspondence: ioannis.mavromatis@digicatapult.org.uk, aftab.khan@toshiba-bril.com.}

\conference{the International Scientific Conference on
Information, Communication and Energy Systems and Technologies (ICEST 2024)---Workshop on Artificial
Intelligence for Sustainable Development (ARISDE 2024), Sozopol, Bulgaria, 1--3 July 2024, which was entitled: Computing Within Limits: An Empirical Study of Energy Consumption in ML Training and Inference} 

\abstract{This study presents an empirical investigation into the energy consumption of Discriminative and Generative AI models within real-world MLOps pipelines. For Discriminative models, we examine various architectures and hyperparameters during training and inference and identify energy-efficient practices. For Generative AI, Large Language Models (LLMs) are assessed, focusing primarily on energy consumption across different model sizes and varying service requests. Our study employs software-based power measurements, ensuring ease of replication across diverse configurations, models, and datasets. We analyse multiple models and hardware setups to uncover correlations among various metrics, identifying key contributors to energy consumption. The results indicate that for Discriminative models, optimising architectures, hyperparameters, and hardware can significantly reduce energy consumption without sacrificing performance. For LLMs, energy efficiency depends on balancing model size, reasoning complexity, and request-handling capacity, as larger models do not necessarily consume more energy when utilisation remains low. This analysis provides practical guidelines for designing green and sustainable ML operations, emphasising energy consumption and carbon footprint reductions while maintaining performance. This paper can serve as a benchmark for accurately estimating total energy use across different types of AI models.}


\usepackage{listings}
\graphicspath{ {./images/} }
\usepackage{siunitx}
\usepackage{nicefrac}
\usepackage[labelformat=simple]{subcaption}

\DeclareCaptionLabelFormat{subcaptionlabel}{\normalfont(\textbf{#2}\normalfont)}
\usepackage{multirow}

\newcolumntype{M}[1]{>{\centering\arraybackslash}m{#1}}
\newcolumntype{R}[1]{>{\raggedleft\arraybackslash}m{#1}}
\def\midtilde@normaltilde{\texttildelow}
\graphicspath{ {./images/} }

\begin{document}



\section{Introduction}
In recent years, Artificial Intelligence (AI) and Machine Learning (ML) have made remarkable strides, transforming numerous sectors. However, their rapid growth has raised concerns about their environmental impact,  with projections indicating that AI/ML pipelines will account for 2\% of global carbon emissions by 2030~\cite{co2Emissions}. The computational demands of training and deploying ML and Deep Learning (DL) models drive significant energy consumption, contributing substantially to carbon emissions. This challenge highlights a pressing question: how can the ML field sustain its advancements while adhering to global sustainability goals?

AI models can be broadly classified into ``Discriminative'' and ``Generative''. Discriminative AI algorithms, such as regression and classification, are used for applications that require high-precision data categorisation and decision-making. Generative AI algorithms focus on creating ``something new'', such as images, text, music and more. Both categories have become increasingly transformative across diverse domains, impacting not only everyday human activities but also specialised industrial applications. For instance, we see Discriminative AI enhancing consumer applications such as shopping with spatial immersion and its synergy with Mixed Reality (MR)~\cite{immersiveMedia}, gaming, entertainment and education~\cite{Moinnereau2022}, and more. Discriminative models are also integral in industry verticals, such as automotive or manufacturing, where they play a critical role in monitoring, automation, and anomaly detection across production lines~\cite{bertolini2021machine}. Such applications highlight ML’s growing presence in key sectors and its ability to address diverse operational needs. 

Generative AI is enabling the creation of high-quality media, text mimicking human-like language and the simulation of complex environments. This branch of AI expands the possibilities for innovation across sectors such as entertainment, healthcare, education, and beyond~\cite{genAI}. Large Language Models (LLMs) exemplify this trend, showcasing remarkable reasoning and understanding abilities that facilitate more interactive and contextually aware user experiences~\cite{li2024large}. Discriminative and Generative models combined can foster AI-native ecosystems such as the emergent intelligent future network~\cite{aiNative6G}, redefining connectivity and the synergy of AI and data exchange. 

However, all the above advancements come at the cost of increased computational requirements: AI/ML models often necessitate large datasets and extensive processing requirements, greatly increasing the energy demands~\cite{mlStrategies}. This is clearly illustrated in the domain of  Generative AI, where datasets and computing resources are vastly larger than conventional Discriminative AI use cases. To tackle the energy demands and mandated Sustainability Development Goals (SDGs) (UN Sustainable Development Goals: \url{https://sdgs.un.org/goals}, accessed on), we see many recent advancements in Green and Sustainable AI practices~\cite{mlStrategies,greenAI}. These practices encompass the efficient use of computational resources and holistic optimisation of ML pipelines. Developing methodologies for energy-efficient ML workflows thus becomes essential for all stakeholders.

Our study builds upon these considerations. We initially discuss the transition from Green Discriminative AI to Green Generative AI. Later, we provide an empirical analysis of energy consumption patterns in both Discriminative and Generative AI applications. For Discriminative AI, we examine both training and inference, analysing various model architectures and hyperparameters to identify areas where energy consumption can be minimised. For Generative AI, we focus on the energy consumption during inference using different tokens and request requirements. Our findings offer key recommendations for reducing energy consumption and propose methods to estimate expected energy use based on various model parameters. Eventually, through analysing the energy costs associated with such tasks, we aim to offer practical guidelines and best practices for researchers and practitioners across the ML Operations (MLOps) lifecycle. While focused on specific tasks, our findings provide generalisable insights for ML practitioners aiming for energy-aware optimisations across diverse use cases.

The remainder of this paper is structured as follows: Sec.~\ref{sec:related_work} presents the SDGs for future systems and recent activities around sustainable Discriminative and Generative AI and discusses their limitations. Green MLOps and the extensions for Generative AI are described in Sec.~\ref{sec:green_mlops}, outlining the energy consumed within an MLOps pipeline. The methodology used for our extensive investigation is illustrated in Sec.~\ref{sec:methodology}. Secs.~\ref{sec:results} and~\ref{sec:discussion} present our results and lessons learned for both large-scale experiments conducted. Finally, the paper is concluded in Sec.~\ref{sec:conclusion}.

\section{Sustainability Goals}\label{sec:related_work}

The United Nations (UN) has recently introduced its 2030 Agenda for Sustainable Development, which outlines 17 SDGs. These SDGs must be taken into account when designing future systems and use cases. Our work aligns closely with the following goals:
\begin{itemize}
    \item \textbf{Goal 9: Industry, Innovation and Infrastructure} - \textit{Build resilient infrastructure, promote inclusive and sustainable industrialisation and foster innovation} - Our work aims to establish a roadmap for developing future MLOps frameworks, fostering innovation and promoting best practices across the technology stack.
    \item \textbf{Goal 10: Reduced Inequalities} - \textit{Reduce inequality within and among countries} - By reducing energy consumption, ML can become more economically viable and sustainable, meeting the 4Cs requirements: Coverage, Capacity, Cost, and Consumption.
    \item \textbf{Goal 12: Responsible Consumption and Production} - \textit{Ensure sustainable consumption and production patterns} - Green ML has the potential to significantly lower reliance on fossil fuels and reduce overall energy consumption.
    \item \textbf{Goal 13: Climate Action} - \textit{Take urgent action to combat climate change and its impacts} - Optimising energy usage across the entire MLOps pipeline can lead to a substantial reduction in carbon emissions.
\end{itemize}

The pursuit of higher accuracy and enriched understanding capabilities leads to larger and more complex models. This trend spans both Discriminative and Generative AI. As AI-native systems grow, their ML pipelines evolve into large-scale operational stages across multiple domains -- from initial data acquisition and pre-processing to model training, deployment, and continuous monitoring. 

Our work addresses the high energy demands associated with both branches of AI. We offer practical approaches and recommendations for creating greener and more sustainable MLOps pipelines, encompassing the entire computing continuum. By providing actionable insights, we aim to promote energy-efficient practices across various use cases and deployment scenarios, ultimately contributing to more sustainable AI-driven systems.

\section{Related Work}\label{subsec:related}
Many studies present concepts and solutions around Green and Sustainable ML. Some notable examples are~\cite{greenAI,mlStrategies,systematicReview}, which focus primarily on Discriminative AI and present statistics on the projected increase in ML's energy consumption over time. Similarly, authors in~\cite{llmSustainabilitySurvey} comment on the economic and sustainability challenges around LLMs and authors of~\cite{mlStrategies} compare transformer models running in Google's data centres. While these works highlight the potential benefits of energy-saving practices (e.g., early exiting, knowledge transfer, etc.), they lack a systematic evaluation of these methods. Our work addresses this gap by conducting an empirical study on real-world hardware.

Traditional energy-saving strategies, such as  pruning~\cite{energyAwarePruning} or quantisation~\cite{energyAwareQuantisation}, have been extensively explored for Discriminative AI in the past. Similar strategies are currently adopted for Generative AI, too, with LLM pruning being proven to be energy-efficient~\cite{llmPruning}. However, usually, such works focus on smaller-scale investigations, impacting the accuracy of a given model. In contrast, contacting a large-scale investigation, we aim to explore ways for energy reductions, examining trade-offs across various configurations and parameters without compromising model accuracy.

Our Generative AI evaluation is primarily focused on the inference of LLMs. Training these large generative models is widely known to be resource-intensive~\cite{cottier2024risingcoststrainingfrontier}; thus requiring substantial energy consumption. Therefore, pre-trained large models are usually used in most real-world generative AI applications. Models such as Meta Llama~\cite{touvron2023llamaopenefficientfoundation} can be either used directly for inference or fine-tuned to meet specific inference needs. Therefore, our investigation will prioritise the inference and how different model sizes can impact the energy consumption of a use case.

The integration of sustainable practices within an ML pipeline is described in~\cite{facebookAI}, published by Meta's AI team. While they tackle the problem systemically and holistically, the individual measurements or models are not detailed. In our work, we analyse well-known models and datasets to enable readers to understand the impact of different hyperparameters, models, and LLM service requests on energy consumption. 

The recent literature includes various relevant studies that evaluate energy consumption with real measurements. The authors of~\cite{mlModelSustainability} focused primarily on shallow single-layer models. Our work will target deeper neural networks to investigate how various hyperparameters affect their training and inference. A work from a few years ago~\cite{Strubell_Ganesh_McCallum_2020} focused on larger transformer-based models but presented only the cost of training and the environmental impact of such models. The model characteristics or hyperparameters exploration were again not considered. More recently,~\cite{llmLlamaPowerCaping} presented an investigation of the Meta's Llama LLM energy consumption across different hardware configurations (GPU sharding, distributed inference, and GPU power capping). This work presented some great insights into hardware domain optimisations. We will follow a similar approach but focus on the trade-offs of the model parameters and the types of requests. Finally,~\cite{llmEnergy} presents a large-scale evaluation of various LLM models and datasets, focusing primarily on how the datasets and the prompt lengths affect energy consumption. Our work aims to extend their findings by investigating the model characteristics that could be optimised for an energy-efficient ML deployment.

\section{From Green MLOps to Green GenOps}\label{sec:green_mlops}
DevOps combines software development and IT operations to shorten the software development cycle and align closely with business goals. It uses integrated tools and automation to streamline software development and delivery. Machine Learning Operations (MLOps) extends DevOps to ML, focusing on the efficient lifecycle management of ML models. It addresses challenges like data management, versioning, and reproducibility while integrating tools for a seamless ML workflow~\cite{li2024adapting}. Most production systems supporting ML-driven applications incorporate an MLOps framework~\cite{mlOpsTaxonomy}. 

LLM Operations (LLMOps), an extension of MLOps, was introduced soon after applications utilising LLMs, such as chatbots, became increasingly popular. This area specifically caters to the nuances of managing LLMs across large-scale systems. However, Generative AI is much bigger than LLMs, incorporating multi-modality across media, data types and systems. Generative Operations (GenOps) or GenAIOps (as it was introduced in~\cite{li2024adapting}) addresses the differences associated with the preparation and handling of vast amounts of unstructured data and the entire spectrum of model management, from pre-training and fine-tuning stages to the intricacies of prompt engineering and the operation of multiple models at scale. In essence, GenOps provides the tools, processes, and practices for orchestrating and automating all stages and functions of the Generative AI model ecosystem, ensuring modularity, scalability, generalisation and compatibility~\cite{li2024adapting}.

The advent of GenOps introduces significant power demands that pose a critical challenge to sustainable and eco-friendly operations. Green MLOps communities have built energy-efficient and cost-effective frameworks for optimising ML and reducing carbon emissions~\cite{greenAI}. However, for GenOps, investigations on energy efficiency are still in their infancy. Building on this foundation, we propose Green GenOps and describe tools and practices that can be used for greener Generative AI operations. The following chapters describe how Green GenOps extends the standard MLOps frameworks and how the energy can be monitored in real time. We also provide insights on energy optimisations during the training and deployment of Discriminative and Generative AI models. Our approach aims to significantly reduce energy consumption while preserving the model's performance and accuracy.

\begin{figure}[t]
    \centering
    \includegraphics[width=0.9\columnwidth]{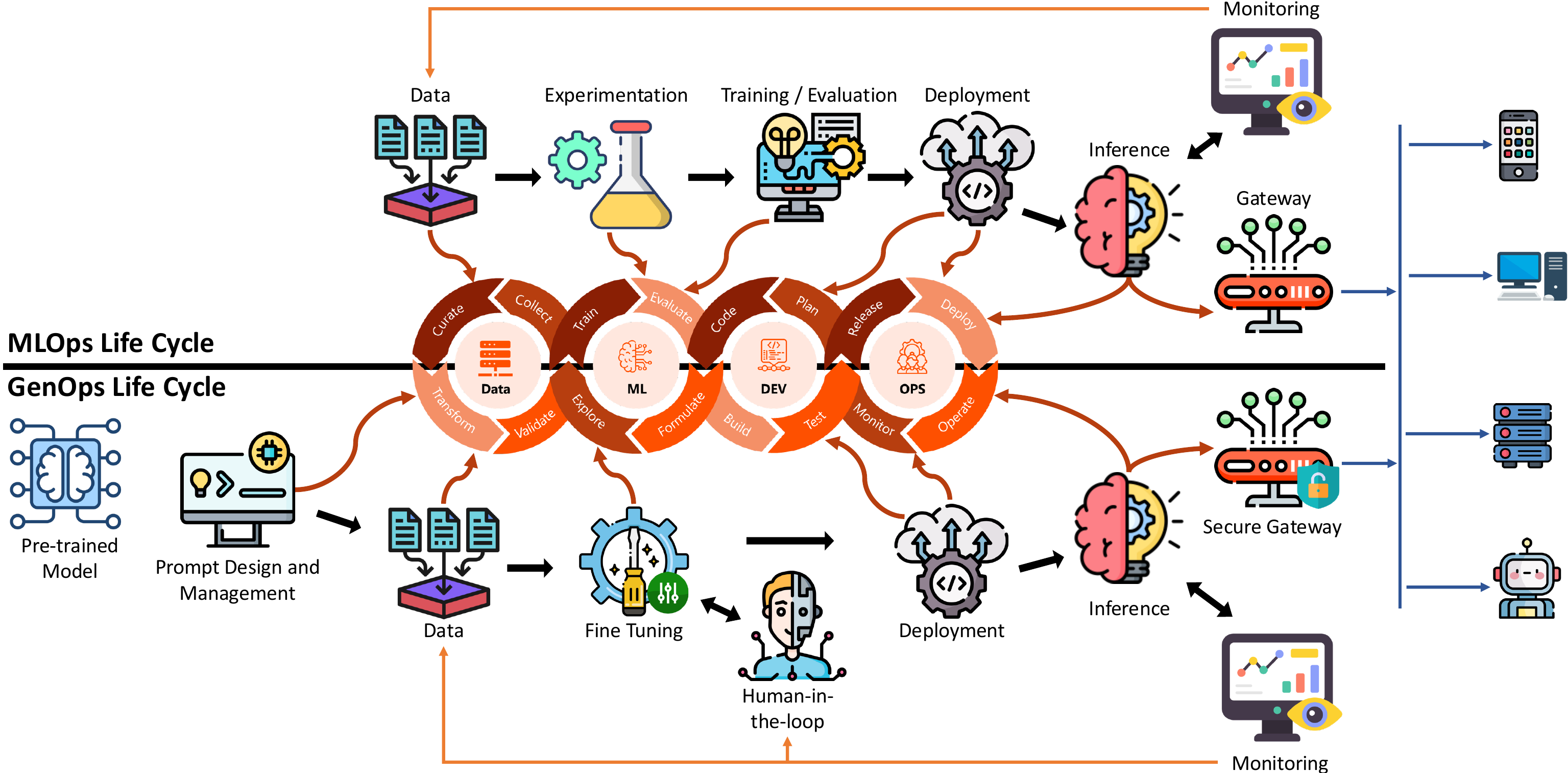}
    \caption{ML model development and deployment phase and the associated MLOps and GenOps life cycles.}
    \label{fig:high_level}
\end{figure}

\subsection{The Transition from MLOps to GenOps}\label{subsec:mlops_to_genops}

MLOps (Fig.~\ref{fig:high_level}-top) typically involves four phases: 1) the \textbf{Data Processing} phase: for collecting, curating, and labelling data, and assigning weights to features, 2) the \textbf{Experimentation} phase: where algorithms, model architectures, and training methods are tested, 3) the \textbf{Training/Evaluation} phase: involves training the selected models on larger, feature-rich datasets, refining the hyperparameters as needed, and finally, 4) the \textbf{Inference} phase: trained models are deployed and take decisions in real-time. All deployed models are usually continuously monitored (part of the \textbf{Inference} phase), measuring their performance and identifying whether a model re-training or model retirement should be triggered. All deployed models are usually packaged as an application (e.g., a microservice) with various exposed interfaces. They are served either running on the service provider's infrastructure, exposed behind a \textbf{Gateway} or shipped to the client to operate in a distributed fashion. 

Moving from MLOps to GenOps (Fig.~\ref{fig:high_level}-bottom), organisations must address, among other challenges, the scale of models (usually requiring specialised infrastructure), the high demands for training and inference, and the unpredictability of the models, i.e., non-deterministic outputs complicate testing and validation. To that extent, as discussed in Sec.~\ref{subsec:related}, foundational \textbf{Pre-trained Models} are usually used to avoid the initial cost required for training (e.g., Meta Llama). A \textbf{Prompt} is a specific input that guides a Generative model to generate a desired output. In GenOps \textbf{Prompt Design and Management} phase is introduced where prompts are created, tested and refined. The finalised and optimised prompts are stored during \textbf{Data Processing} phase and can usually be shared among multiple projects. While foundational models are good at generalising, it is a common practice to have a \textbf{Model Fine-tuning} phase, where a model is specialised on specific tasks or domains, using curated datasets and prompts. The supervised fine-tuning usually involves a \textbf{Reinforcement Learning from Human Feedback (RLHF)} phase, where a human-in-the-loop helps fine-tune the model’s behaviour over time. When a model is marked as ready (adequately fine-tuned), it is deployed at the service provider's infrastructure and is exposed to the end-users via standardised interfaces. The exposed model is usually accompanied by a \textbf{Secure Gateway}, where guardrails and filters are applied to both prompts and model outputs to prevent harmful responses. Finally, as before, the Generative model is continuously monitored to identify drift or harmful/malicious operations. 

\subsection{Energy Consumption in MLOps and GenOps and Sustainability}\label{subsec:energy_consumption}

From the above, GenOps can be seen as the evolution of MLOps, taking into account all the intricacies of Generative AI models and excluding unnecessary operations (e,g., the training). Recent applications and deployments are seen to merge traditional MLOps approaches with GenOps pipelines while using multiple Discriminative and Generative AI models in synergy~\cite{hybridML,hybridML2}. It is seen that various models can be combined for hybrid (Discriminative and Generative) inferences or that Discriminative models are used for the optimisation and monitoring of GenOps pipelines. This leads to increasingly complex systems that need to manage, orchestrate, monitor, train and infer on multiple models with different architectural specifications while handling a vast number of requests. The complexity of such a system collectively increases the energy consumption and the environmental impact even more. 

For a traditional MLOps pipeline, training, experimenting, and inferring account for a significant portion of the energy consumed~\cite{Strubell_Ganesh_McCallum_2020}. Facebook's AI research team~\cite{facebookAI} indicates that inference requires more compute cycles than training, having a split of $10\%:20\%:70\%$ between \textbf{Experimentation}, \textbf{Training/Evaluation} and \textbf{Inference}, respectively. While we could not find any investigations that report the energy consumption across the different phases of GenOps, we believe a similar split is very likely. It will not be surprising if the inference consumes an even more significant portion at the end. Considering the energy distribution across the entire MLOps pipeline, again~\cite{facebookAI} reports that it is roughly $31\%:29\%:40\%$ for the \textbf{Data}, \textbf{Experimentation/Training/Evaluation}, and \textbf{Inference} phases. Overall, poor optimisation strategies, inadequate hyperparameter tuning and poor neural network management can vastly increase energy consumption. As described in~\cite{Strubell_Ganesh_McCallum_2020}, this could increase the energy consumption by up to $\times2000$ times for Natural Language Model (NLP) models and up to $\times3000$ for a transformer-based NLP. Data management and pipeline optimisations are considered out-of-scope for this work, so we focus on phases that require training or inference.

GenOps, extends traditional application architectures in various ways. For example, while microservices form the fundamental operation unit in DevOps and MLOps, Generative AI introduces the concept of AI agents~\cite{genAIAgents}. These agents are discrete, reusable, and decoupled units designed to handle specific tasks. GenOps also incorporates non-deterministic reasoning loops, breaking tasks into smaller, domain-specific, iterative steps that reduce computational overhead. New model definitions manage multi-modal context and systems under a single operational framework, one can streamline workflows and resource allocation. Finally, efficient prompt design and refining, prompt caching, and reusing optimised prompts are central to reducing computational overhead. These elements are critical for Green GenOps and necessitate specialised operations for energy-efficient management of GenOps workflows.

In the above-described systems, various works have proposed solutions on the energy-efficient prompt design~\cite{llmEnergy}, energy-aware hardware and resource optimisation~\cite{llmLlamaPowerCaping}, pruning techniques~\cite{llmPruning} that reduce the total energy consumption and more. However, none of these works focused on how model characteristics and number of requests impact the energy consumption of an MLOps or GenOps pipeline. This will be the gap addressed by this paper. For Discriminative AI, we will investigate both training and inference and how parameters such as the model size, the batch size, the time required for training and inference, etc., affect the energy consumption. Similarly, for Generative AI, we focus exclusively on the inference stage and examine how varying per-second request rates impact the energy consumption of different sizes of LLMs. Overall, our findings and recommendations will target ML practitioners who aim to build Green GenOps pipelines at scale that combine the operation of both Discriminative and Generative models within the same unified framework.

\section{Methodology}\label{sec:methodology}
In order to calculate the total energy consumption for an experiment, we need to measure the absolute power at frequent intervals. The time required for each experiment is also essential. Hardware statistics like the utilisation of resources and the model characteristics should also be captured as part of our experimentation and correlated with the model characteristics and hyperparameters. More information about the framework implemented for the Discriminative AI evaluation can be found at~\cite{energy_ml}.

\subsection{Gathering Software-Based Energy Consumption Data}\label{subsec:energyAPIs}
Monitoring energy consumption can be accomplished using hardware or software tools. Hardware-based methods offer high precision~\cite{physicalMeter}, but they face challenges in synchronisation and control~\cite{hardwareSync}, particularly for brief measurements, such as evaluating a shallow neural network. These methods often require external clocks and expensive equipment, making them less accessible to many ML practitioners. Our investigation adopts a software-based approach to measure energy consumption. This approach not only reduces costs and complexity but also ensures greater consistency and scalability. Additionally, it enables parallel evaluations across multiple devices and allows us to measure the power consumption consistently for both Discriminative and Generative models. 

Software-based energy measurement typically employs one of two approaches. The first estimates power consumption using a hardware component's Thermal Design Power (TDP) and its utilisation, assuming a linear relationship between the two. TDP, measured in Watts (\SI{}{\watt}), represents the maximum power consumption under full theoretical load. However, this method oversimplifies the relationship between power consumption and utilization~\cite{LIN20211045}, as modern hardware dynamically adjusts the frequency and can deactivate entire cores to conserve energy. A more sophisticated approach derives power consumption from the hardware’s capacitance ($C$), voltage ($V$), and frequency ($f$), using the formula $P = \nicefrac{1}{2} \ C V^2 f$. While this method provides a more accurate representation, obtaining precise values for these parameters across all hardware components is often impractical

As a workaround, manufacturers provide access to energy data through Model Specific Registers (MSRs), such as Nvidia's Management Library (NVML) for GPUs and Intel's Running Average Power Limit (RAPL) for CPUs and DRAM usage. These methods are reliable with a reported variance of about $\pm \SI{5}{\watt}$ in absolute values while maintaining consistent trends in relative measurements~\cite{Nvidia2016,erqTradeOffVideoCodecs}. For consumer CPUs where MSRs do not provide DRAM measurements, DRAM energy consumption is approximated using the formula $P_{\mathrm{DRAM}} = \sum N_{\mathrm{DIMM}} \times P_{\mathrm{DIMM}}$, where $N_{\mathrm{DIMM}}$ is the number of DIMMs and $P_{\mathrm{DIMM}} = \nicefrac{1}{2} \ C V^2 f$. The operational $V$ and $f$ are accessible from the OS, and $C$ is fixed for all our experiments. This equation is a good approximation as voltage variations during DRAM operations are almost negligible, and operational frequency does not change over time~\cite{dramPowerConsumption}.

Our experimental methodology is as follows. We trigger the execution of the energy measuring toolkit and the training/inference application for a given scenario at the same time. At the end of the experiment, the training/inference application triggers the termination of the energy measuring toolkit, and the toolkit stores the results for post-processing. This process is iterated across all scenarios multiple times, and our results are averaged out across all runs.

\subsection{Calculating Energy Usage in Machine Learning Processes}\label{subsec:energyMeasurements}
Our investigation focuses on either training or inference sessions. To measure the energy consumption we define two metrics, i.e., $E_{\mathrm{tr}}$, which is the total energy consumed during one training session (i.e., for a given model and dataset, with a pre-defined set of hyperparameters and a fixed number of epochs), and $E_{\mathrm{in}}$, which is the total energy during inference (i.e., for a given model and dataset, inferring across all samples with a given batch size). They are as follows:
\begin{equation}\label{eq:training}
    E_\mathrm{tr}=\int^{T_\mathrm{tr}}_{t=0} P_\mathrm{tr}(t) \,dt-\int^{T_\mathrm{idle}}_{t=0} P_\mathrm{idle}(t) \,dt
\end{equation}
\begin{equation}\label{eq:inference}
    E_\mathrm{in}=\int^{T_\mathrm{in}}_{t=0} P_\mathrm{in}(t) \,dt-\int^{T_\mathrm{idle}}_{t=0} P_\mathrm{idle}(t) \,dt
\end{equation}
where $T_\mathrm{tr}$ and $T_\mathrm{in}$ are the training and inference times, $T_\mathrm{idle}$ is a hardcoded time interval used for the idle experiment, and $P_\mathrm{tr}$, $P_\mathrm{in}$ and $P_{\mathrm{idle}}$ are the power measurements during training, inference and when the system is idle. 

While Discriminative AI models usually run on a single machine, it is not uncommon for Generative AI models to be split across multiple GPU servers or multiple GPUs within the same server. Moreover, many enterprise servers utilise multiple CPU sockets and packages. Therefore, power consumption calculations should take that into consideration and as seen later, for our calculations we consider the sum of the power consumption of all hardware components involved. We capture the power consumption at frequent intervals $\Delta t$. Denoting $t_i$ as the $i$-th time interval, the power $P(t_i)$ (this could be either for training or inference) is:
\begin{equation}
P(t_i) = \sum^{N_\mathrm{CPU}}_{k=1} P_\mathrm{CPU_k}(t_i) + \sum^{N_\mathrm{GPU}}_{k=1} P_\mathrm{GPU_k}(t_i) + \sum^{N_\mathrm{DRAM}}_{k=1}P_\mathrm{DRAM_k}(t_i)
\end{equation}
where $P_\mathrm{CPU}$, $P_\mathrm{GPU}$ and $P_\mathrm{DRAM}$ are the power consumption, taken in real-time for the CPU socket (CPU package), GPU socket and DRAM DIMM, respectively. The energy within $i$-th interval can be calculated as the $E(t_i) = P(t_i)\, \Delta t$. Based on that, the Eqs.~\eqref{eq:training} and~\eqref{eq:inference} can be approximated with the cumulative sum of all intervals, i.e.:
\begin{equation}\label{eq:training_sum}
    E_\mathrm{tr}=\sum^{N_\mathrm{tr}}_{i=0} P_\mathrm{tr}(t_i) \,\Delta t-\sum^{N_\mathrm{idle}}_{t=0} P_\mathrm{idle}(t_i) \,\Delta t
\end{equation}
\begin{equation}\label{eq:inference_sum}
    E_\mathrm{in}=\sum^{N_\mathrm{in}}_{t=0} P_\mathrm{in}(t_i) \,\Delta t-\sum^{N_\mathrm{idle}}_{t=0} P_\mathrm{idle}(t_i) \,\Delta t
\end{equation}
where $N_\mathrm{tr}$, $N_\mathrm{in}$  and $N_\mathrm{idle}$ are the total number of intervals during training, inference, or idle, respectively. As discussed, data exchange and processing, even though they play a significant role in the energy consumed, will not be considered.

\begin{table}[t]
\fontsize{9pt}{9pt}\selectfont
    \centering
    \caption{Hardware Configurations (HCs). In brackets is the TDP for each hardware component.}
    \begin{tabular}{lllll@{}}
    \toprule
    & HC-1 & HC-2 & HC-3 & HC-4\ \\
    \midrule
    CPU$^*$ & i7-8700K (\SI{95}{\watt}) & i9-11900KF (\SI{125}{\watt}) & i5-12500 (\SI{65}{\watt}) & Xeon 8480+ (\SI{350}{\watt}) \\ \midrule
    \multirow{2}{*}{DRAM} & $4\times\SI{16}{\giga\byte}$ DDR4 & $4\times\SI{32}{\giga\byte}$ DDR4 & $2\times\SI{16}{\giga\byte}$ DDR5 & $16\times\SI{64}{\giga\byte}$ DDR5 \\
     & \SI{3600}{\mega\hertz} & \SI{3200}{\mega\hertz} & \SI{3200}{\mega\hertz} & \SI{2200}{\mega\hertz} \\ \midrule
    \multirow{2}{*}{GPU$^+$} & RTX 3080 (\SI{320}{\watt}) & RTX 3090 (\SI{350}{\watt}) & RTX A2000 (\SI{70}{\watt}) & $2\times$H100 ($2\times$\SI{300}{\watt}) \\
     & \SI{10}{\giga\byte} & \SI{24}{\giga\byte} & \SI{12}{\giga\byte} & $2\times$\SI{80}{\giga\byte} \\
    \bottomrule
    \end{tabular}
    \newline
    $^*$Intel Core, $^+$Nvidia driver v530.30.02, CUDA v12.1
    \label{tab:pcs}
\end{table}

\subsection{Hardware Stats and Model Characteristics}
In Table~\ref{tab:pcs}, we list all the hardware configurations used for our experiments. As all configurations use Intel CPU sockets and Nvidia GPUs, we utilised RAPL or NVML libraries, respectively, for all measurements. Moreover, we collect various utilisation and thermal values during execution. The NVML library provides the GPU (and its VRAM) utilisation. For the CPU, the utilisation metrics were directly collected from the OS as a function of each CPU core. The CPU utilisation is calculated as the average utilisation at a given time between all cores. Similarly, DRAM's utilisation was also captured directly from the OS.

As described earlier, our evaluation aims to identify patterns and model characteristics that can affect total energy consumption. To achieve some consistency across the generative and discriminative experiments, we identified various model metrics that could be measured for both. These include the \textit{model size}, the~number of \textit{total and trainable parameters}, and~
\textit{multiply--accumulate operation (MAC)}. Moreover, for the discriminative AI use-case, we also captured the \textit{buffer size} and the floating-point operations per second (FLOPS) for the Generative AI experiment. 

The model size, measured in bytes (\SI{}{\byte}), is calculated when the model is decompressed and loaded in the VRAM. It includes both the parameters and buffers and represents the overall footprint of the model in memory. Particularly for Generative AI models, measuring their size instead is critical as it is the major limiting factor on LLM deployment. Depending on the load, the computational power of the GPU may not be the bottleneck towards higher throughput, but the model size may be.

The total number of parameters and the trainable parameters are key indicators of a model's complexity. Trainable parameters differ when certain layers in the model are frozen (i.e., not updated during training). Generally, a larger number of parameters implies a more complex model, which may achieve higher accuracy but at the cost of increased computational resources and memory usage. This added complexity can lead to longer training times and may necessitate more powerful hardware.

The buffer size represents additional data structures used for storing intermediate outputs and constants that remain unchanged during training, such as batch normalization parameters. While these do not directly contribute to the model's learning capacity, they significantly affect the overall memory footprint. A large buffer size can result in inefficiencies, particularly in systems with limited memory.

FLOPs and MACs are metrics commonly used to calculate the computational complexity of deep neural networks. FLOPs refer to the number of arithmetic operations—addition, subtraction, multiplication, and division—performed on floating-point numbers. These operations are central to many mathematical computations in ML, including matrix multiplications, activations, and gradient calculations. FLOPs are commonly used to quantify the computational cost or complexity of a model or a specific operation within it. This metric estimates the total arithmetic operations required, making it particularly useful for assessing computational efficiency. By measuring FLOPs, researchers and practitioners can better understand and compare the resource demands of different models or configurations.

Finally, MACs specifically count the number of operations where two numbers are multiplied, and the result is added to an accumulator. This operation is fundamental to numerous linear algebra tasks, including matrix multiplications, convolutions, and dot products. MACs provide a more targeted measure of computational complexity, particularly in models that heavily rely on linear algebra operations, such as Convolutional Neural Networks (CNNs). By focusing on these critical operations, MACs offer a practical metric for assessing the computational demands of such models.

For our investigation, these model characteristics -- whether analysed independently or in combination -- are assessed to explore their impact on total energy consumption. These parameters are calculated when the model is loaded onto the GPU before the execution of each experiment.

\section{Results}\label{sec:results}

For our investigation, we performed two sets of experiments, one focusing on Discriminative AI and another on Generative AI. The following sections describe our power consumption measurements and our initial observations, and Sec.~\ref{sec:discussion} delves into our findings and how these could be applied in an ML deployment. Finally, each section describes the evaluation metrics for the Discriminative and Generative AI experiments used in this study.

\begin{table}[t]
    \centering
    \caption{Model Parameters for Discriminative AI experiments}
    \begin{tabular}{r|l}
        \textbf{Hyperparameter} & \textbf{Value} \\
        \toprule
        Batch Size & 128 \\
        Learning Rate & 0.001 \\
        Optimizer & Stochastic Gradient Descent \\
        Loss Function & Categorical Cross-Entropy \\
        Weight Decay & $5 \times 10^{-4}$ \\
        \bottomrule
    \end{tabular}
    \label{tab:model_setup}
\end{table}

\subsection{Discriminative AI models}
We investigated Discriminative AI with a simple image classification task, an application very common in hand gesture detection, interactive educational games, etc.~\cite{imageClassification, educationalGames}. This application was chosen due to the abundance of models and datasets available in the literature. The selected model architectures span various sizes and types. We chose: SimpleDLA, DPN (26), DenseNet (121), EfficientNet (B0), GoogLeNet, LeNet, MobileNet, MobileNetV2, PNASNet, PreActResNet (18), RegNet (X\_200MF), ResNet (18),  ResNeXt (29\_2x64d), SENet (18), ShuffleNetV2, and VGG (16), to analyse the behaviours of different models. The number in the parenthesis specifies the model variant chosen for our experiment. All experiments were conducted with the same hyperparameters (batch size of $128$, learning rate $0.001$,  stochastic gradient descent optimiser, categorical cross-entropy loss and weight decay $5\times10^{-4}$). To maintain consistency across runs, we also fixed the random seed. Our model parameters are also summarised in Tab.~\ref{tab:model_setup}. Variations in the hyperparameters used across the different experiments are described in each section.

\begin{figure}[t]
    \centering
    \begin{subfigure}{0.99\columnwidth}
        \centering
        \includegraphics[width=\linewidth]{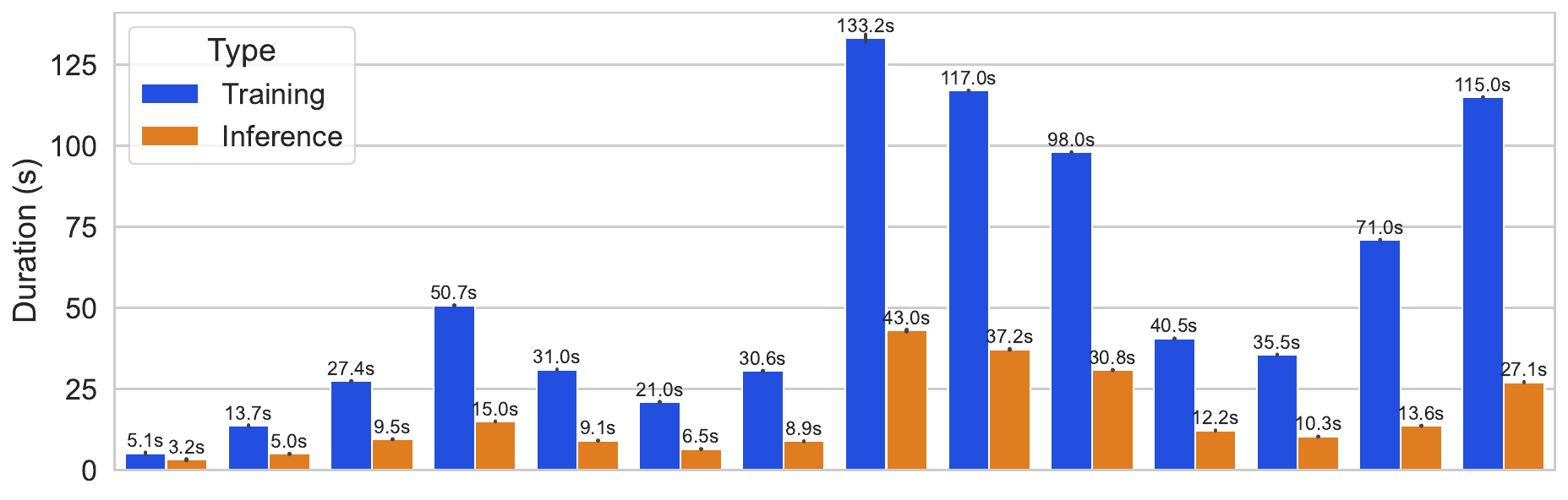}
        \caption{\centering Duration for HC-3.}
        \label{fig:duration_a2000}
    \end{subfigure}%
    \\\vspace{3mm}
    \begin{subfigure}{0.99\columnwidth}
        \centering
        \includegraphics[width=\linewidth]{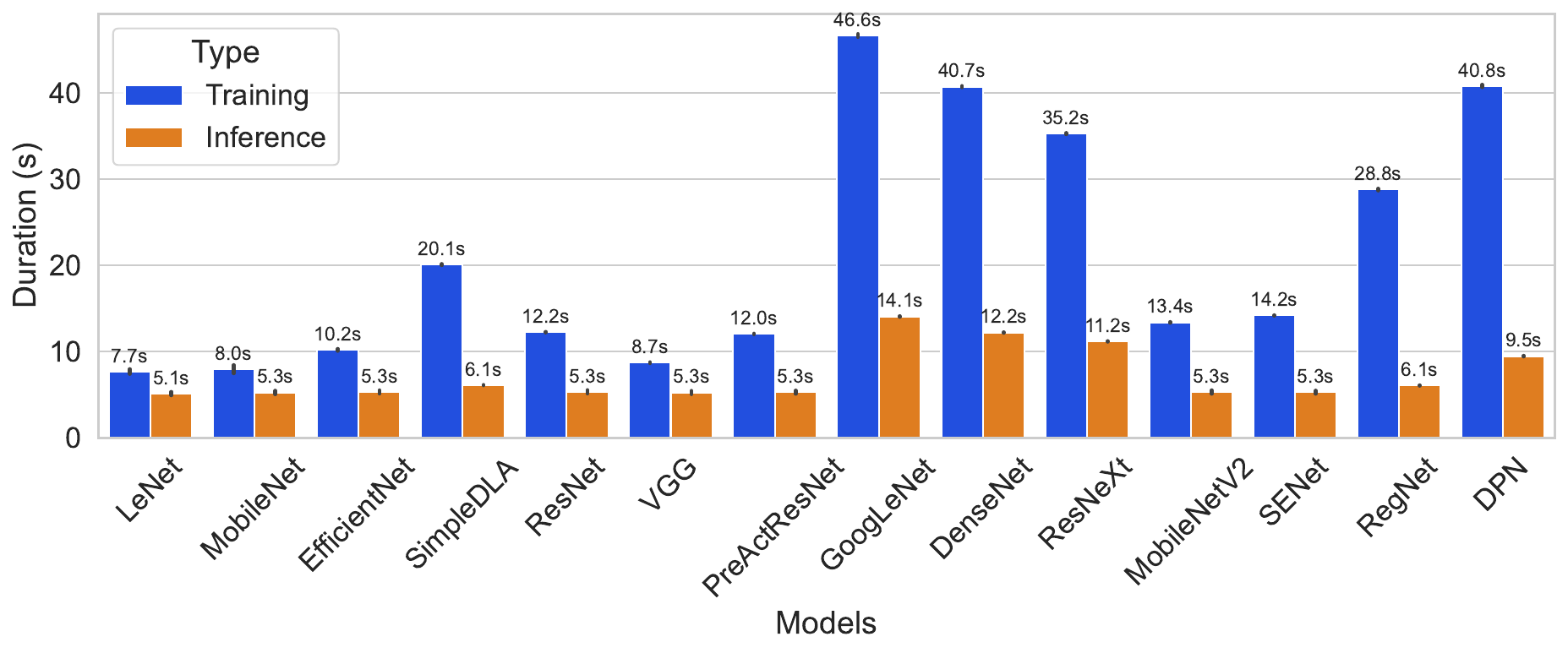}
        \caption{\centering Duration for HC-2.}
        \label{fig:duration_3090}
    \end{subfigure}
    \caption{Training and inference duration (for $50$k samples).}
    \label{fig:combined_duration}
\end{figure}

The experiments are based on the first three different Hardware Configurations (HCs) summarised in Table~\ref{tab:pcs}. These three HCs provide varied environments to explore and identify their differences or similarities and the correlations (Pearson $r$ and Spearman $\rho$) of the different model parameters. We used the CIFAR-10 dataset~\cite{cifar10}, which consists of $60000$ $32\times32$ RGB colour images across 10 classes equally split per class, e.g., aeroplane, bird, cat, dog, etc. ($6000$ images per class). All images were normalised per channel using the CIFAR-10 training set statistics (mean = (0.4914, 0.4822, 0.4465), std = (0.2023, 0.1994, 0.2010)), ensuring each input has approximately zero mean and unit variance. CIFAR-10 was chosen due to its popularity in benchmarking a wide range of image classification models, from lightweight networks to deeper convolutional architectures. The split between the training and testing set is $50000:10000$. For evaluation, the testing set was replicated fivefold (i.e., to $50$k samples) to ensure consistency between training and inference samples.

\subsubsection{Initial Statistics}\label{subsec:init_statistics}
The accuracy achieved by most models was between $87\%-91\%$ after 100 epochs. As expected, the shallower LeNet underperformed, reaching only around $68\%$, while MobileNet and EfficientNet achieved $81\%$ and $83\%$, respectively. The training and inference durations (one epoch of training and inference on $50$k samples) are shown in Fig.~\ref{fig:combined_duration}. For most models, training takes approximately three times longer than inference due to the computational overhead of backpropagation and parameter updates ($r \approx 0.9$ across all models and HCs). However, models like DPN and RegNet deviate from this trend.

\begin{figure}[t]
    \centering
    \begin{subfigure}{0.99\columnwidth}
        \centering
        \includegraphics[width=\linewidth]{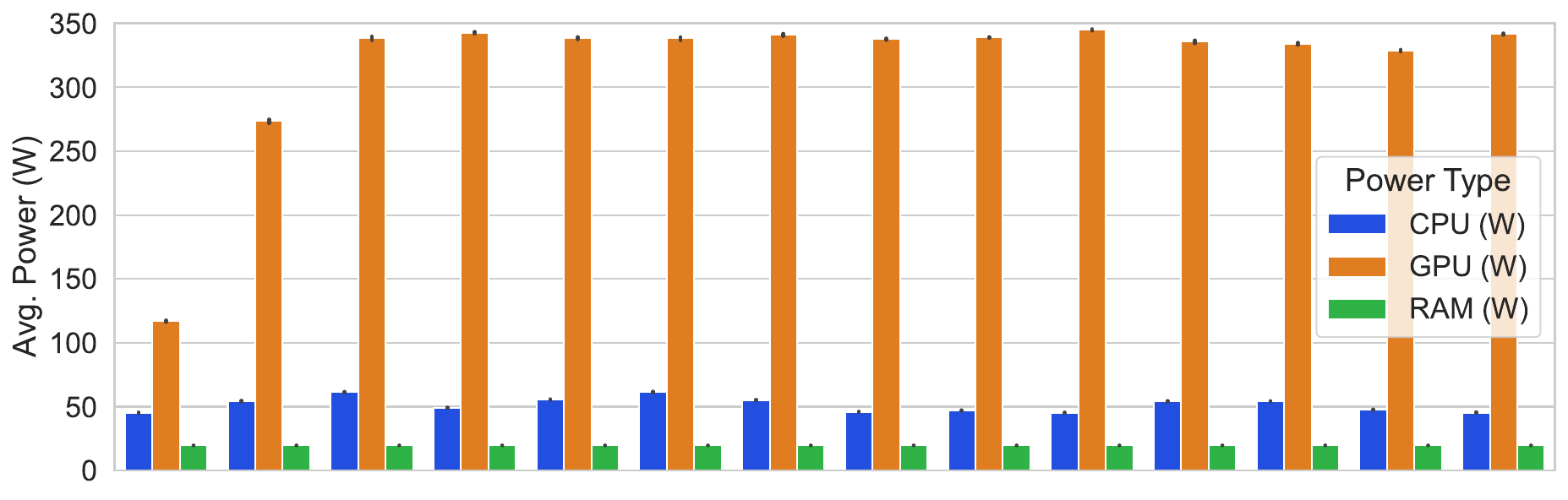}
        \caption{\centering Power usage by model (training).}
        \label{fig:power_training}
    \end{subfigure}%
    \\\vspace{3mm}
    \begin{subfigure}{0.99\columnwidth}
        \centering
        \includegraphics[width=\linewidth]{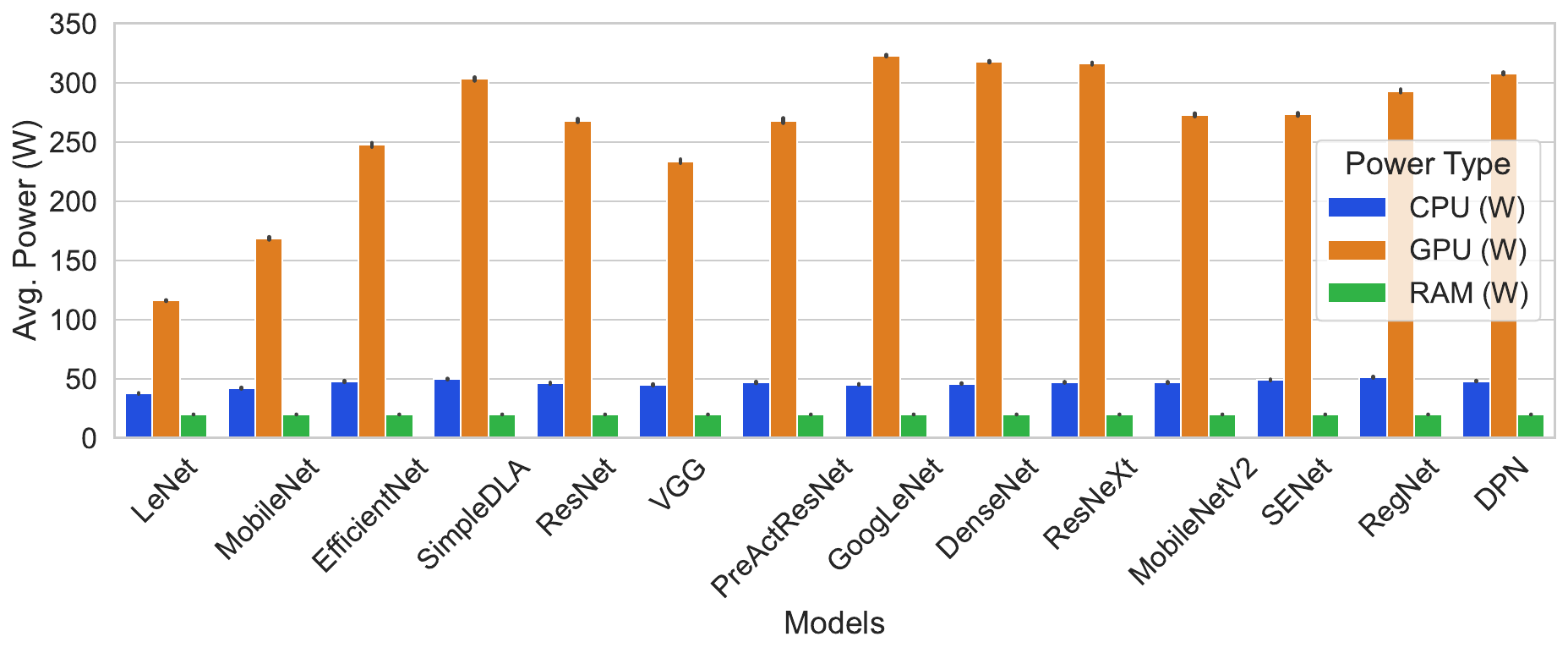}
        \caption{\centering Power usage by model (inference).}
        \label{fig:power_testing}
    \end{subfigure}
    \caption{Average power usage with HC-2.}
    \label{fig:combined_power}
\end{figure}

Significant differences were observed across hardware configurations (HCs) for the same models. For instance, PreActResNet at HC-2 (Fig.~\ref{fig:duration_3090}) requires about $5\mathrm{x}$ more time to train or infer compared to LeNet, but at HC-3 (Fig.~\ref{fig:duration_a2000}), that difference increases to $26\mathrm{x}$.  Interestingly, during training, the relative time differences between models remained consistent, but during inference, smaller models on a more powerful GPU (HC-2) processed the same number of samples in nearly identical durations, regardless of model size. Given that inference largely determines energy consumption (as discussed in Sec.~\ref{subsec:energy_consumption}), models that achieve similar accuracy but infer faster offer significant long-term energy savings, even if their training times are longer. For example, VGG and ResNet deliver comparable accuracy to DenseNet or DPN but consume only a fraction of the energy, making them more suitable for prolonged use.

\subsubsection{Power Consumption Measurements - Discriminative AI}
Fig.~\ref{fig:combined_power} illustrates the average power consumed for HC-2 for training and inference. For larger models, the GPU operates close to its TDP, as shown in Fig.~\ref{fig:power_training}. As expected, CPU and DRAM, being underutilised, exhibit roughly equal and not significantly high average power consumption across all models. However, this differs from the inference, as depicted in Fig.~\ref{fig:power_testing}. Many models operate $\geq 30\%$ below the GPU's TDP (e.g., VGG), whereas CPU and DRAM follow the same trends as with the training. The same applies across all HCs, with the difference being more prominent for HC-1 and less prominent for HC-3.

\begin{figure}[t]
    \centering
    \begin{subfigure}{0.99\columnwidth}
        \centering
        \includegraphics[width=\linewidth]{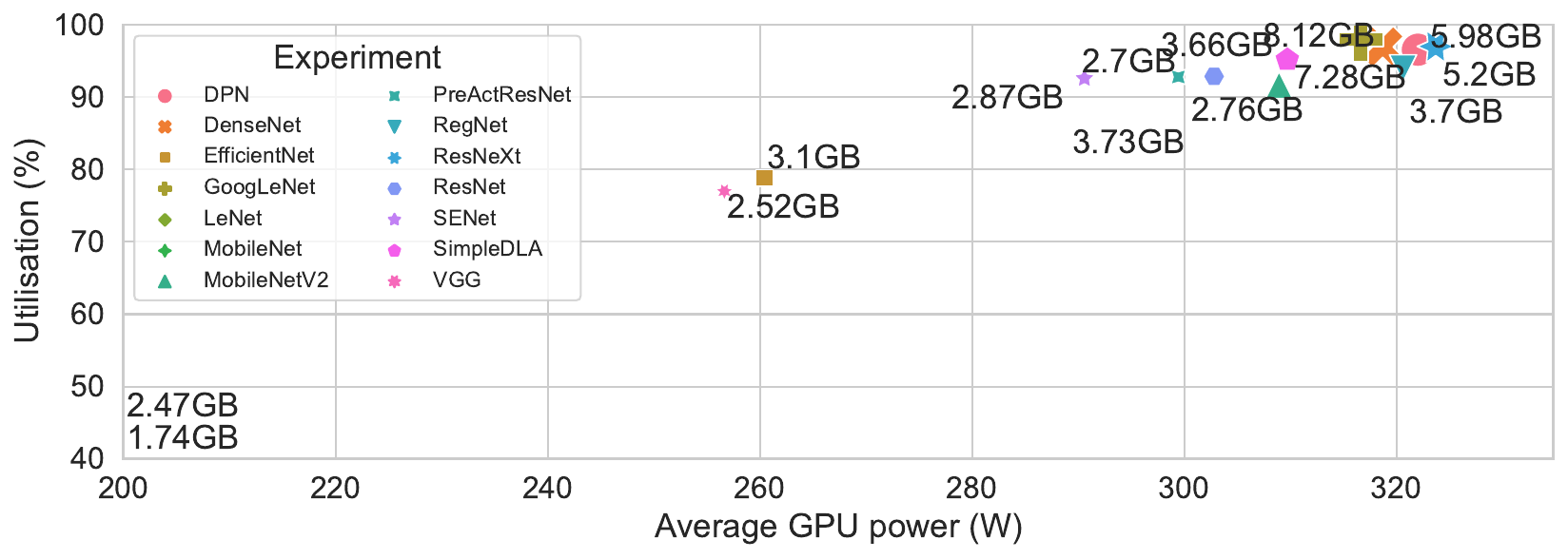}
        \caption{\centering Power usage by model (training).}
        \label{fig:gpu_ram_training}
    \end{subfigure}%
    \\\vspace{3mm}
    \begin{subfigure}{0.99\columnwidth}
        \centering
        \includegraphics[width=\linewidth]{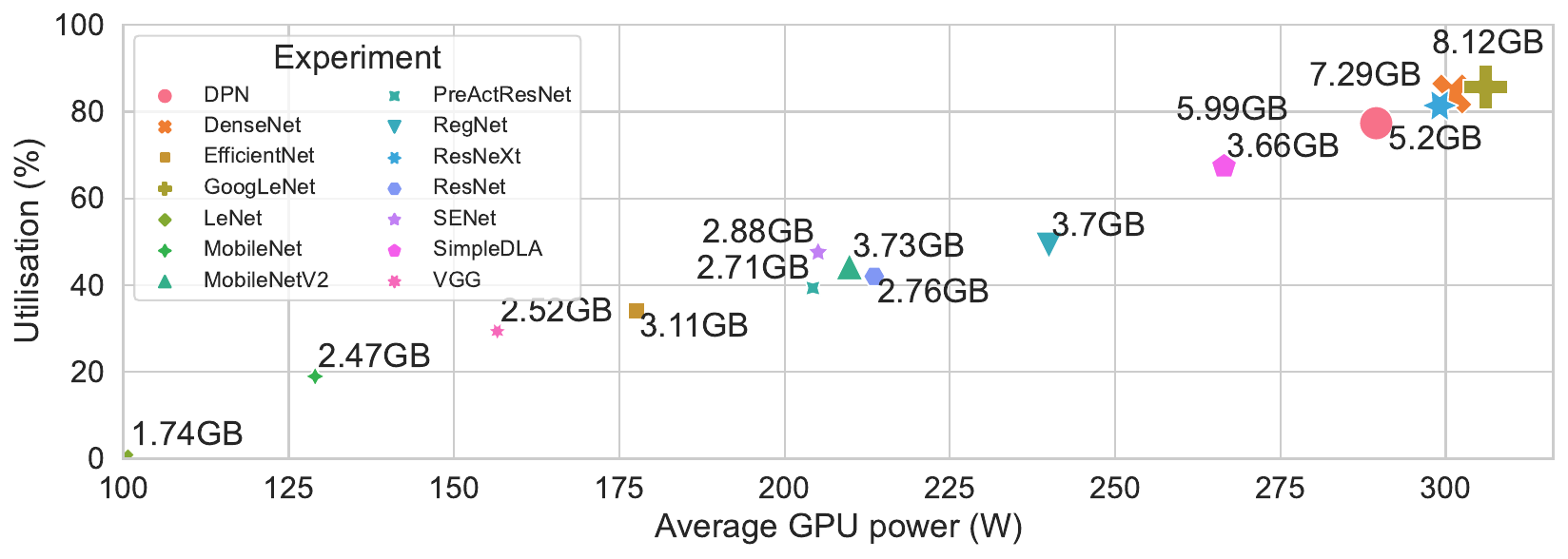}
        \caption{\centering Power usage by model (inference).}
        \label{fig:gpu_ram_testing}
    \end{subfigure}
    \caption{Utilisation and power consumption (considering the GPU RAM usage) - HC-1.}
    \label{fig:gpu_ram}
\end{figure}

Since CPU and DRAM usage remains relatively constant across different models, we compare the power consumption with the GPU (VRAM and processing resources) utilisation (Fig.~\ref{fig:gpu_ram}). A larger GPU VRAM use generally corresponds to higher utilisation and greater power consumption, a trend that is more noticeable during inference. Our results indicate a strong correlation between utilisation and power consumption. Although, this correlation holds up to a certain threshold (e.g., $\rho \approx 0.81$ for HC-3, $\rho \approx 0.55$ for HC-2). Beyond a power draw of \raisebox{-0.6ex}{\~{}}\SI{300}{\watt}, further increases in the GPU utilisation did not result in increases in the power consumption. This is clearer in Fig.~\ref{fig:gpu_ram_training}, where most models push the GPU to operate close to its TDP. Our findings in this study align with our previous work~\cite{frost}.

\begin{figure}[t]
    \centering
    \includegraphics[width=0.95\columnwidth]{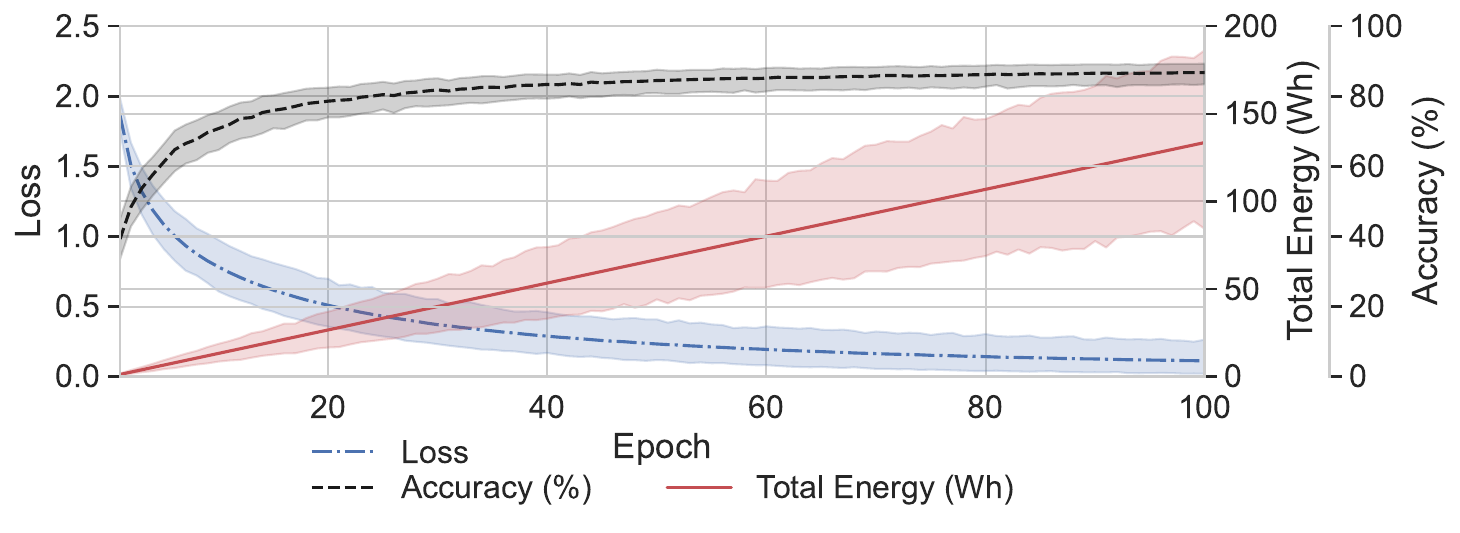}
    \caption{Loss, energy and accuracy per epoch, averaged across all models - the shaded areas show the range of values - HC-3.}
    \label{fig:loss_power}
\end{figure}

Our investigation reveals a strong linear relationship between time and energy consumption, with $r = 0.99$ (e.g., per training epoch or fixed number of samples during inference). While comparing the model loss, accuracy, and total energy accumulated as the number of epochs increases (average across all models while training -- Fig.~\ref{fig:loss_power}), even though there is no correlation between accuracy and total energy consumed, as the number of epochs increases, the range of values observed for the energy, is greater (relatively) to the accuracy, thus replacing a model can significantly benefit the energy consumption with no significant cost in the accuracy.

\begin{figure}[t]
    \centering
    \begin{subfigure}{0.99\columnwidth}
        \centering
        \includegraphics[width=\linewidth]{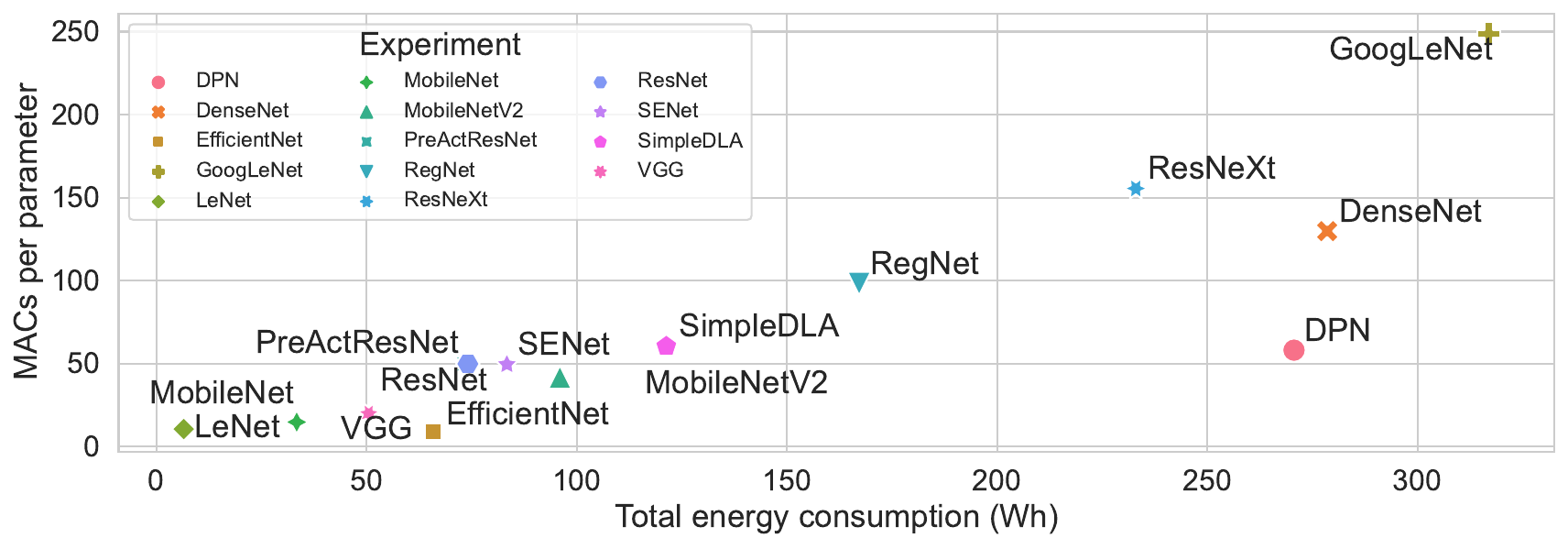}
        \caption{\centering During the training phase.}
        \label{fig:macs_training}
    \end{subfigure}
    \\\vspace{3mm}
    \begin{subfigure}{0.99\columnwidth}
        \centering
        \includegraphics[width=\linewidth]{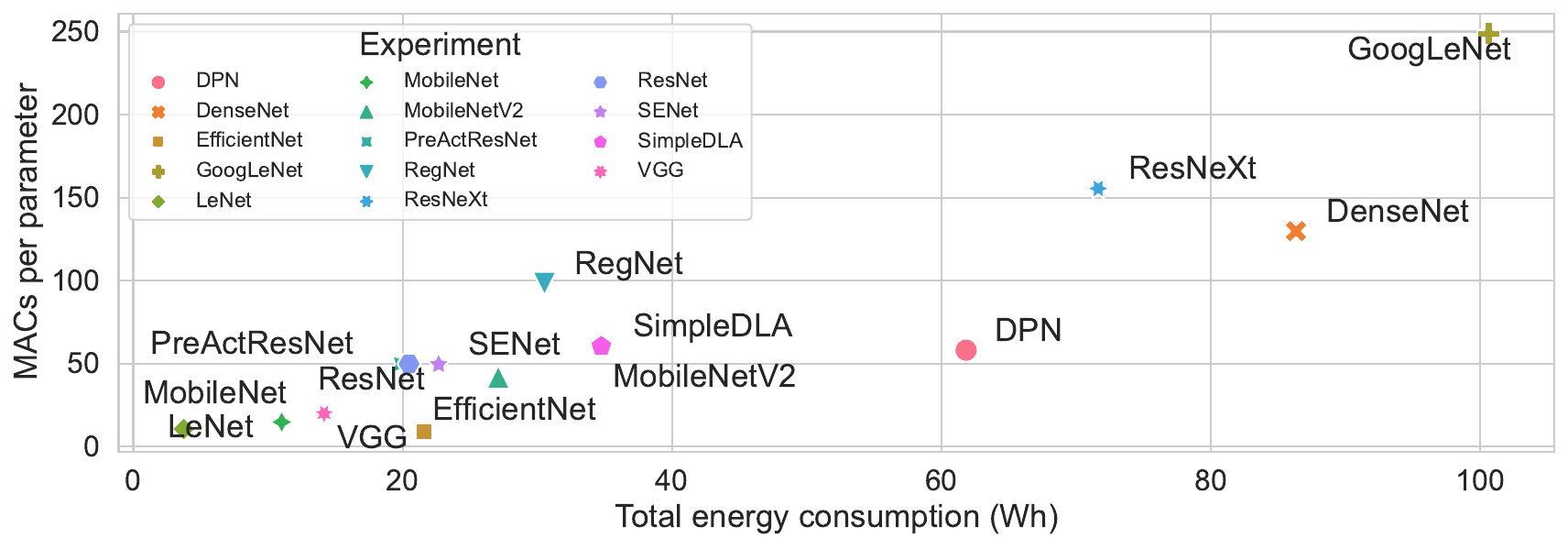}
        \caption{\centering During the inference phase.}
        \label{fig:macs_testing}
    \end{subfigure}
    \caption{Total energy consumption as a function of the MACs per parameter - HC-3.}
    \label{fig:macs_parameters}
\end{figure}

MAC is usually a standard metric commonly used to assess the complexity of a model and its expected energy consumption. When we compare the MACs of different models in relation to their total energy usage, we find a strong correlation between them, with $\rho \approx 0.8$ across all HCs. However, our analysis indicates that combining MACs with the model parameters (Fig.~\ref{fig:macs_parameters}) provides a more representative metric. For both training (Fig.~\ref{fig:macs_training}) and inference (Fig.~\ref{fig:macs_testing}), we see a strong correlation across them ($\rho \approx 0.9$ across all HCs).

\begin{figure}[t]
    \centering
    \begin{subfigure}{0.99\columnwidth}
        \centering
        \includegraphics[width=\linewidth]{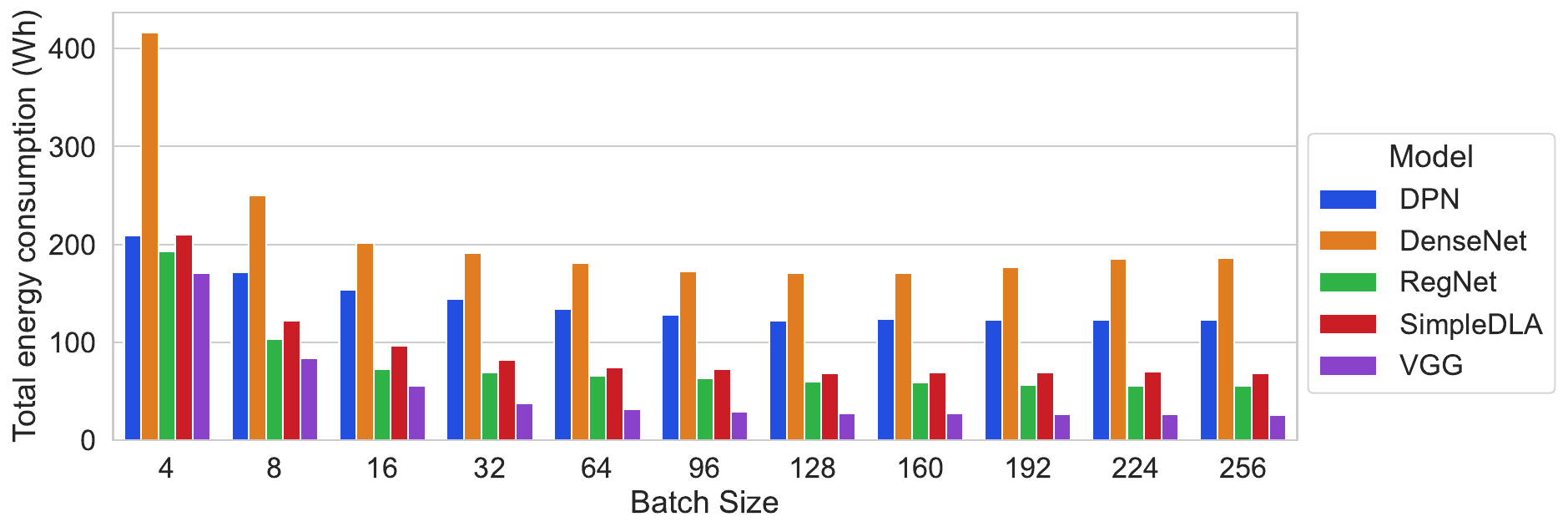}
        \caption{\centering Batch size and total energy consumption.}
        \label{fig:batch_energy}
    \end{subfigure}
    \\\vspace{3mm}
    \begin{subfigure}{0.99\columnwidth}
        \centering
        \includegraphics[width=\linewidth]{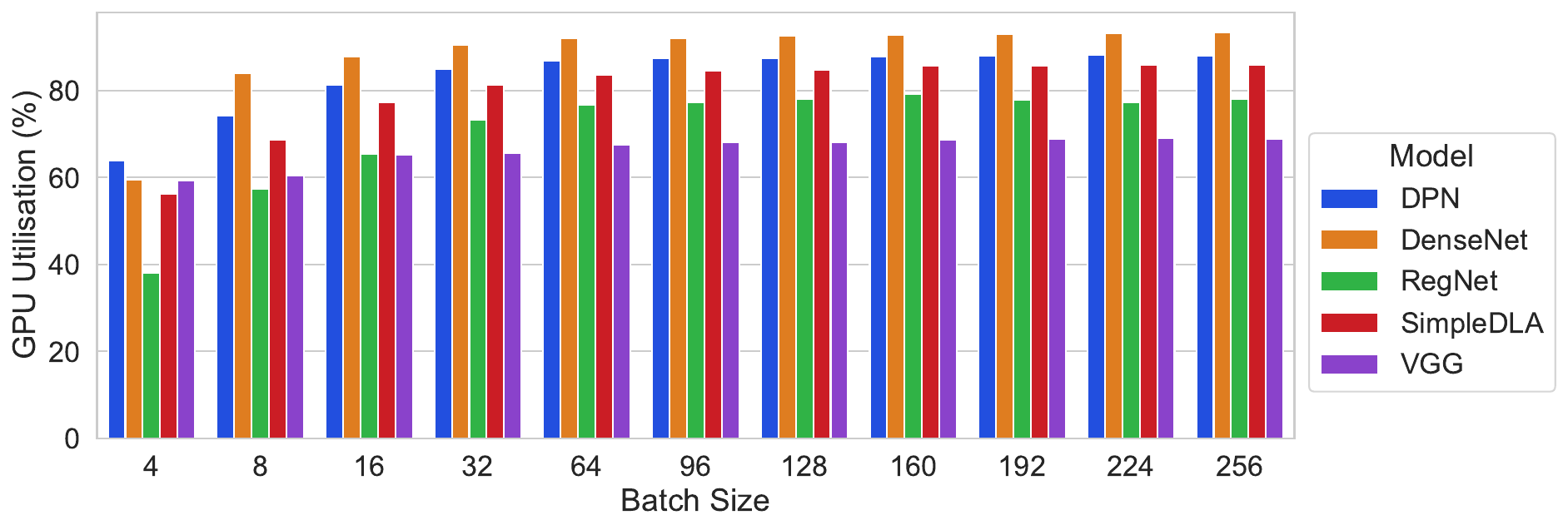}
        \caption{\centering Batch size and GPU utilisation.}
        \label{fig:batch_utilisation}
    \end{subfigure}
    \caption{Effect of batch size on total energy consumption and GPU utilisation - HC-2.}
    \label{fig:batch}
\end{figure}

Finally, when comparing different batch sizes for training and inference (Fig.~\ref{fig:batch}), we find that smaller batch sizes tend to increase power consumption (Fig.~\ref{fig:batch_energy}). This increase directly correlates with the GPU utilisation for each model (Fig.~\ref{fig:batch_utilisation}). For every HC, an optimal batch size exists that minimises power consumption; any further increase in the batch size does not yield additional improvements. Importantly, as smaller batch sizes achieve higher accuracy~\cite{batchSizeComparison}, this indicates a tradeoff between the accuracy and the energy consumption that requires further exploration.

\subsubsection{Total and GPU-only Energy Consumption and Correlation Metrics}\label{subsec:computation_metrics}

As discussed earlier, inference is expected to be the most energy-consuming phase of an ML pipeline due to the volume of samples being inferred in a real-world system. We, therefore, present in Table~\ref{tab:spearman_correlations} the correlation of various metrics with the total energy consumption, focusing on the inference phase. Investigating the same values for the GPU energy consumption in isolation, we identified no significant difference between them; therefore, we do not include them in the paper.

We devised nine metrics to provide insights into the model's performance, energy consumption and resource utilisation. These were: 
\begin{enumerate}
    \item \textbf{\texttt{macs\_param}}: Calculated as the ratio of MACs to trainable parameters -- evaluates the computational efficiency of the model architecture (also seen in Fig.~\ref{fig:macs_parameters})
    \item \textbf{\texttt{work\_done}}: Defined as the trainable parameters processed per second -- assesses computational throughput and resource utilisation
    \item \textbf{\texttt{overall\_efficiency}}: The ratio of the accuracy multiplied by the \textbf{\texttt{work\_done}} over the system's utilisation
    \item \textbf{\texttt{energy\_per\_sample}}: Represents the total average energy consumption for one sample of inference
    \item \textbf{\texttt{parameters}}: The total trainable parameters, a key indicator of model complexity and context for other metrics
    \item \textbf{\texttt{work\_per\_unit\_power}}: Calculated as \textbf{\texttt{work\_done}} divided by the observed power for a given batch of samples, quantifying energy efficiency
    \item \textbf{\texttt{energy\_scaling\_factor}}: The ratio of the total power (CPU, GPU and RAM) to model parameters
    \item \textbf{\texttt{gpu\_energy\_scaling\_factor}}: Similar to \textbf{\texttt{energy\_scaling\_factor}} but focused on just the GPU's absolute power consumption, -- both show how energy consumption scales with model complexity
    \item \textbf{\texttt{model\_size\_to\_ram}}: Compares model size to memory usage, aiding in optimising memory efficiency for resource-limited systems
\end{enumerate}

\begin{table}[t]
    \centering
    \caption{Spearman Correlations of the total energy consumptions and various metrics.}
    \begin{tabular}{lccc}
        \toprule
        \textbf{Metric}                & \textbf{HC-1} & \textbf{HC-2} & \textbf{HC-3} \\ 
        \midrule
        energy\_per\_sample           & 1.000000      & 1.000000      & 1.000000      \\ 
        macs\_param                   & 0.902342      & 0.915271      & 0.852587      \\ 
        model\_size\_to\_ram          & 0.521170      & 0.212621      & 0.457989      \\ 
        overall\_efficiency           & -0.439481     & -0.340809     & -0.592853      \\ 
        work\_per\_unit\_power        & -0.311792     & -0.388402     & -0.691502     \\ 
        gpu\_energy\_scaling\_factor  & 0.229738      & 0.196945      & 0.390812      \\ 
        energy\_scaling\_factor       & -0.039773     & -0.106112     & -0.109445      \\ 
        parameters                    & 0.200979      & 0.212926      & 0.142314      \\ 
        work\_done                    & 0.021486      & -0.052404     & -0.387764     \\ 
        \bottomrule
    \end{tabular}
    \label{tab:spearman_correlations}
\end{table}

We see that the temporal correlation between the energy consumption and a single sample's inference makes the \textbf{\texttt{energy\_per\_sample}} a highly reliable energy predictor regardless of the hardware. Similarly, the strong correlation of \textbf{\texttt{macs\_param}} across different hardware configurations indicates that computational efficiency is a strong and consistent factor in energy consumption. From the \textbf{\texttt{work\_done}}, it is indicated that just the ``throughput'' of a pair ''ML model/hardware configuration'' is not directly tied to the energy consumption. However, the moderate negative correlations of the \textbf{\texttt{overall\_efficiency}} for HC-1 and HC-2 (with the mid-tier hardware showing a better correlation) and the strong correlation for HC-3 indicate that particularly for energy-efficient hardware configurations, there is a higher correlation between the system's efficiency and the energy consumption and short-living experiments can be used for extrapolating the expected energy over longer periods. 

The negative values of \textbf{\texttt{work\_per\_unit\_power}} indicate that higher efficiency is associated with lower energy consumption. However, the top-tier hardware (HC-2) does show a higher correlation compared to the mid-tier one (HC-1) (something that is not the case in the \textbf{\texttt{overall\_efficiency}}), indicating that hardware architecture differences have to be considered for long-term deployments. Also, the higher value in the low-tier hardware (HC-3) indicates an energy-optimised hardware and an energy-performance tradeoff that can be considered when orchestrating model deployments across heterogeneous hardware configurations.

The moderate correlation of the \textbf{\texttt{model\_size\_to\_ram}} with the energy consumption shows that the model size compared to the total VRAM available plays a role but is not a dominant factor. This is intuitive, as other factors (e.g., computation) likely overshadow memory usage in energy scaling. Finally, for the \textbf{\texttt{energy\_scaling\_factor}}, the \textbf{\texttt{gpu\_energy\_scaling\_factor}}, and the \textbf{\texttt{parameters}}, we observe a weak correlation. The number of parameters is not a strong determinant of energy use, with model architectural factors such as the MACs playing a more significant role. Similarly, from the \textbf{\texttt{energy\_scaling\_factor}}, the \textbf{\texttt{gpu\_energy\_scaling\_factor}}, we see that the absolute power consumption and, to that extent, the total energy consumed does not scale with the number of parameters.

\begin{figure}[t]
    \centering
    \includegraphics[width=1\columnwidth]{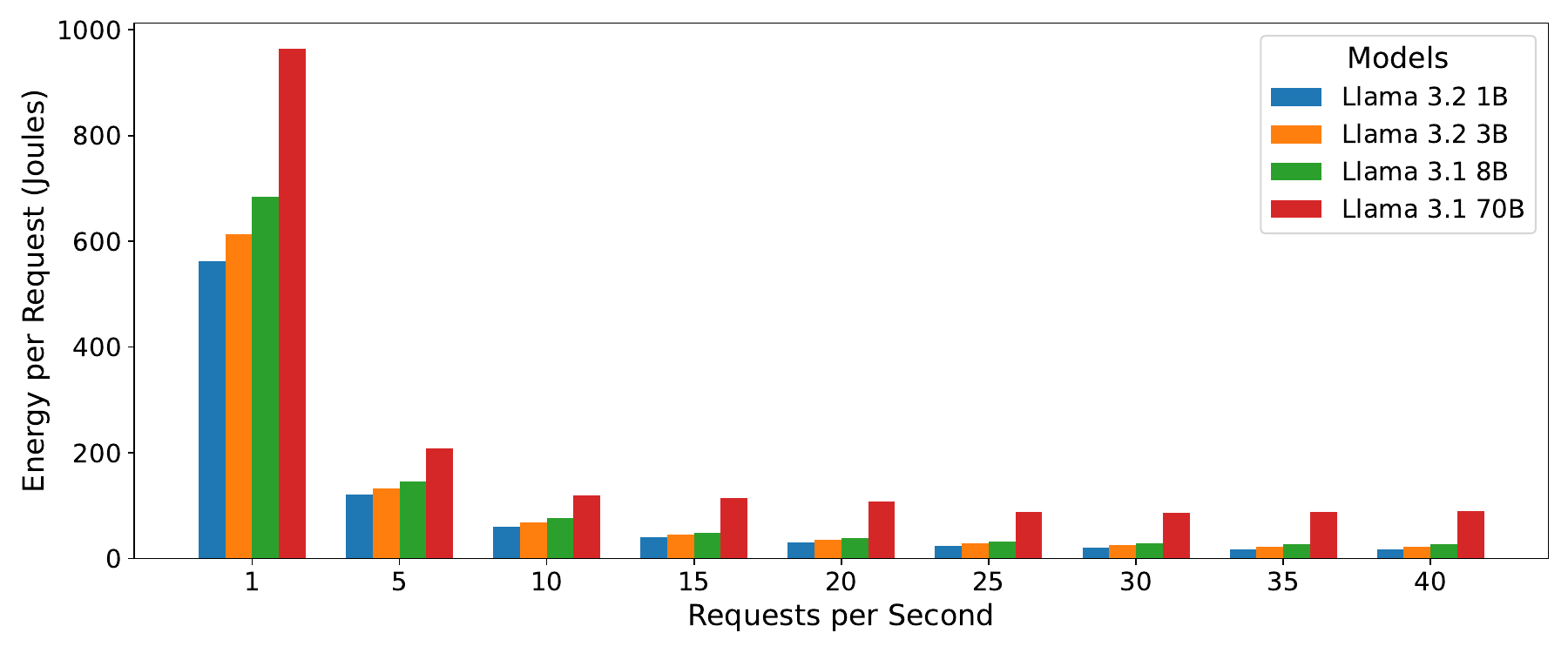}
    \caption{Total energy consumption per request as a function of the number of RPS.}
    \label{fig:total_energy_per_request}
\end{figure}

\begin{figure}[t]
    \centering
    \includegraphics[width=1\columnwidth]{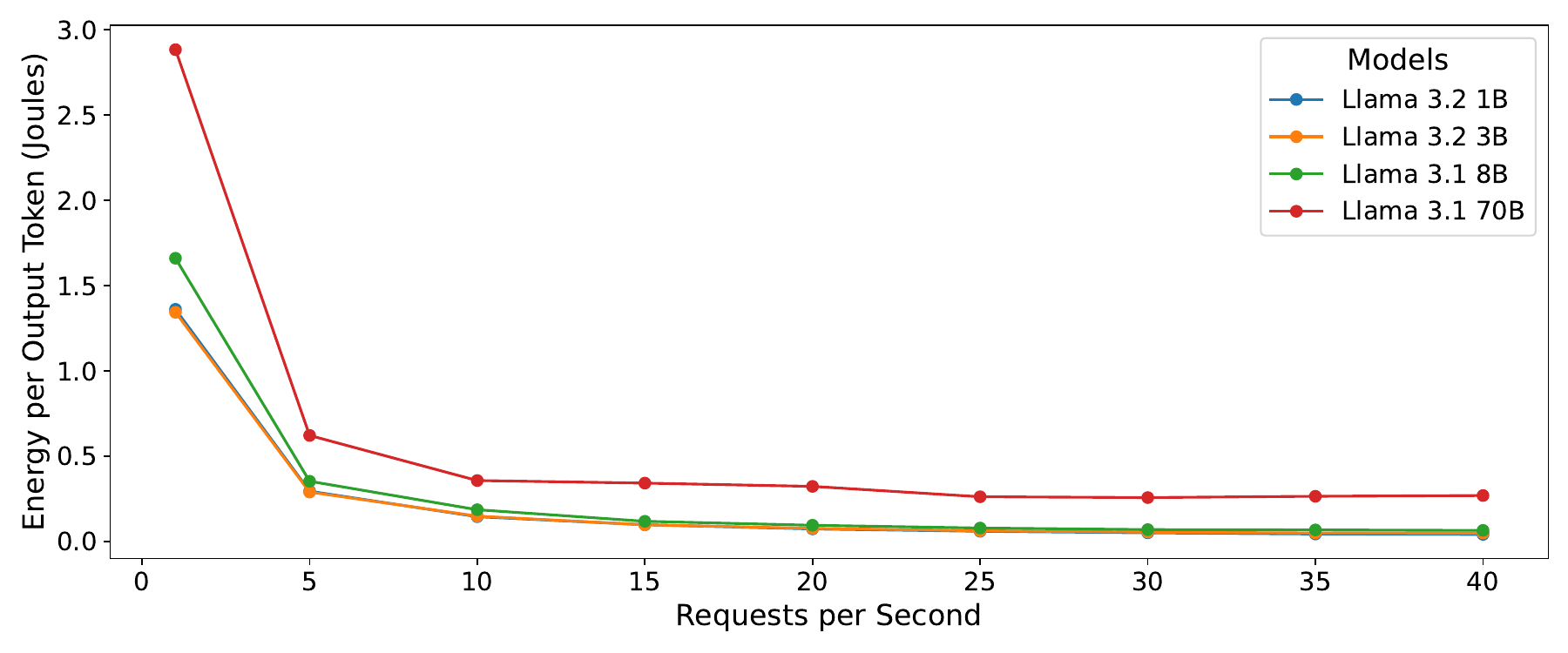}
    \caption{Energy per output token as a function of RPS.}
    \label{fig:energy_per_output_token}
\end{figure}

\begin{figure}[t]
    \centering
    \includegraphics[width=1\columnwidth]{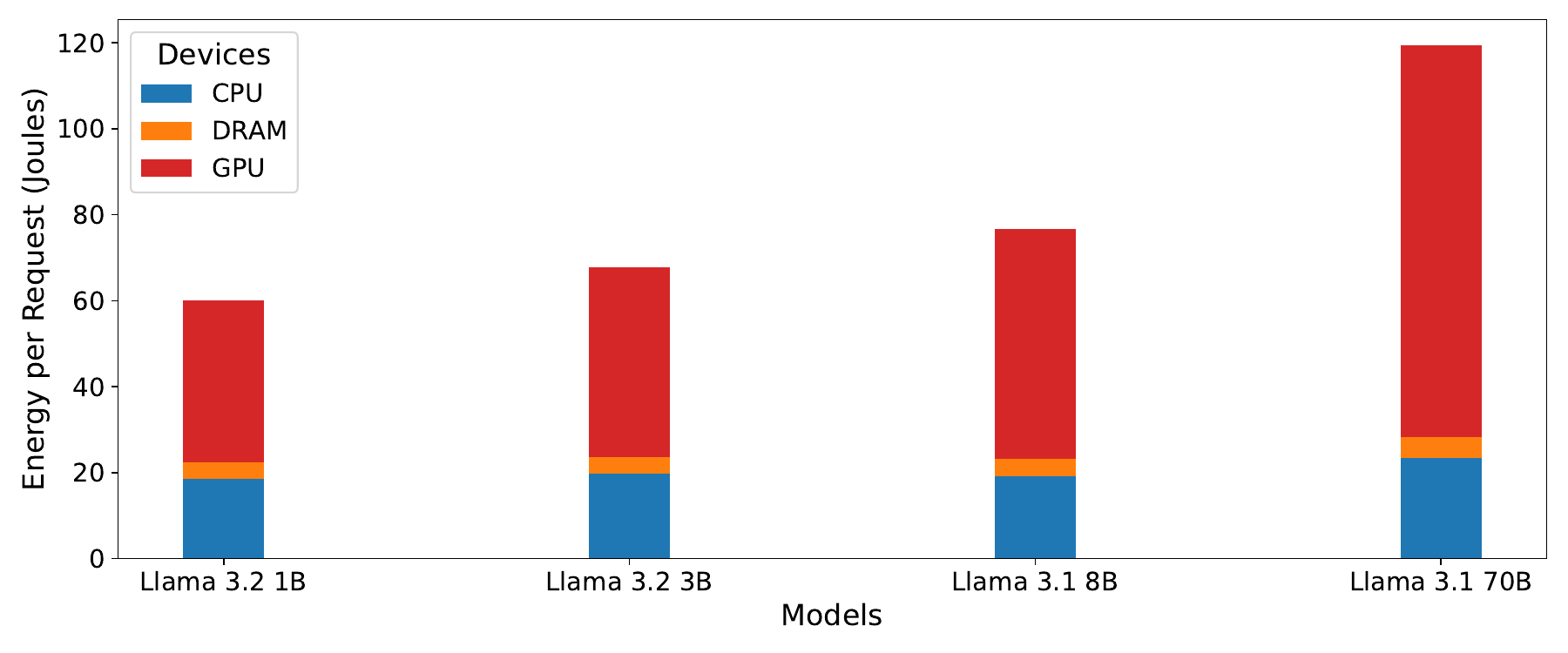}
    \caption{Per-device energy consumption per request at 10 RPS.}
    \label{fig:per_device_energy}
\end{figure}

\subsection{Generative AI models}

To analyse the energy consumption of Generative AI models in the context of LLM inference, we focus on tasks involving real-time, high-frequency interactions, such as those encountered in chatbot platforms. We conducted our experiments on a high-performance hardware configuration (HC-4 in Table~\ref{tab:pcs}) consisting of two Nvidia H100 GPUs, an Xeon 8480+ CPU, and substantial DRAM capacity. This setup allows us to efficiently manage the computational demands of inference tasks at various request rates, simulating real-world applications where LLMs respond to multiple concurrent users.

\begin{table}[t]
    \centering
    \caption{Model Parameters for Generative AI experiments}
    \begin{tabular}{r|l}
        \textbf{Hyperparameter} & \textbf{Value} \\
        \toprule
        Temperature & 0 \\
        Top-p & 1 \\
        Top-k & -1 \\
        Min-p & 0 \\
        Detokenisation & True \\
        \bottomrule
    \end{tabular}
    \label{tab:model_setup_llm}
\end{table}

We picked different-sized models to provide reference points for different applications. These models are part of the Meta Llama family of models, particularly the 1 and 3 billion parameter models from the 3.2 generation and the 8 and 70 billion parameter models from the 3.1 generation. These models' weights are quantised for their inference to 8-bit floating point numbers. Their activation functions remain non-quantised. These models were deployed to a vLLM inference endpoint~\cite{kwon2023efficient}; a state-of-the-art LLM inference engine allowing multi-threaded Generative AI operation (i.e., multiple concurrent conversations being answered simultaneously). The LLM hyperparameters fixed across all experiments were the: temperature $0$, top-p, $1$, top-k $-1$, min-p $0$ and detokenisation ``true''. These are also summarised in Table~\ref{tab:model_setup_llm}. We measured energy usage while varying the Requests Per Second (RPS), a critical parameter directly impacting the model's computational load and energy requirements. Specifically, we employed the Chatbot Arena~\cite{zheng2023judging} dataset that contains real human queries to chatbots (as per the examples found in Table~\ref{tab:chatbot_arena_questions}) to replicate high-traffic conditions, where user interactions necessitate continuous and rapid LLM responses. By simulating different RPS levels, we aimed to capture the energy footprint of Generative AI under various operational scenarios, providing insights into sustainable deployment practices.

\begin{table}[t]
    \centering
    \caption{Sample questions from the Chatbot Arena dataset}
    \begin{tabular}{ll}
    \hline
        \textbf{ID} & \textbf{Question} \\
        \hline
        1 & What is the difference between OpenCL and CUDA? \\
        2 & Why did my parent not invite me to their wedding? \\
        3 & Fuji vs. Nikon, which is better? \\
        \hline
        \end{tabular}
    \label{tab:chatbot_arena_questions}
\end{table}

In the following sections, we present detailed power consumption measurements for the LLMs under different RPS settings, identify the primary factors contributing to energy usage, and discuss strategies for optimizing energy efficiency during Generative AI model inference.

\subsubsection{Power Consumption Measurements - Generative AI}
Power consumption data was collected by measuring the energy per request across different RPS settings to capture the responsiveness and efficiency of each model configuration under variable loads. The results are displayed in Fig.~\ref{fig:total_energy_per_request}, which shows the energy per request across the models tested. The data provides insight into the relationship between RPS and energy consumption, indicating that as RPS increases, the per-request energy cost initially decreases due to more efficient utilisation of GPU resources. However, the energy cost per request stabilizes or slightly increases beyond a certain threshold due to resource saturation. The resource saturation of the concurrent processing threads available for each model saturate at 40 RPS for the 1 and 3 billion parameter models, 35 RPS for the 8 billion parameter model, and 10 RPS for the 70 billion parameter model.

Fig.~\ref{fig:energy_per_output_token} illustrates the energy consumption per output token across various RPS settings. The graph shows that smaller models maintain lower energy costs per token at higher RPS values, reflecting their suitability for high-throughput scenarios. Conversely, larger models like the 70B configuration exhibit significantly higher energy consumption per token, particularly at lower RPS values, due to the computational intensity required.

Fig.~\ref{fig:per_device_energy} presents the per-device energy consumption per request for the tested models operating at 10 RPS. The results reveal that CPU and DRAM consumption remain relatively consistent across the models, only slightly increasing as the model size scales. In contrast, GPU consumption significantly rises with larger models, reflecting their increased utilisation of GPU compute resources. Specifically, the GPU energy consumption for the 70B model is nearly three times that of the 1B model. For smaller models like 1B, 3B, and 8B, which do not fully utilise the available GPU compute resources, the observed energy consumption increases incrementally. However, the transition to the 70B model results in a dramatic surge in GPU energy consumption, underscoring the exponential growth in computational demand as model size increases. This highlights the need for targeted GPU workload optimisation to effectively manage energy efficiency for larger models.

\begin{figure}[t]
    \centering
    \includegraphics[width=1\columnwidth]{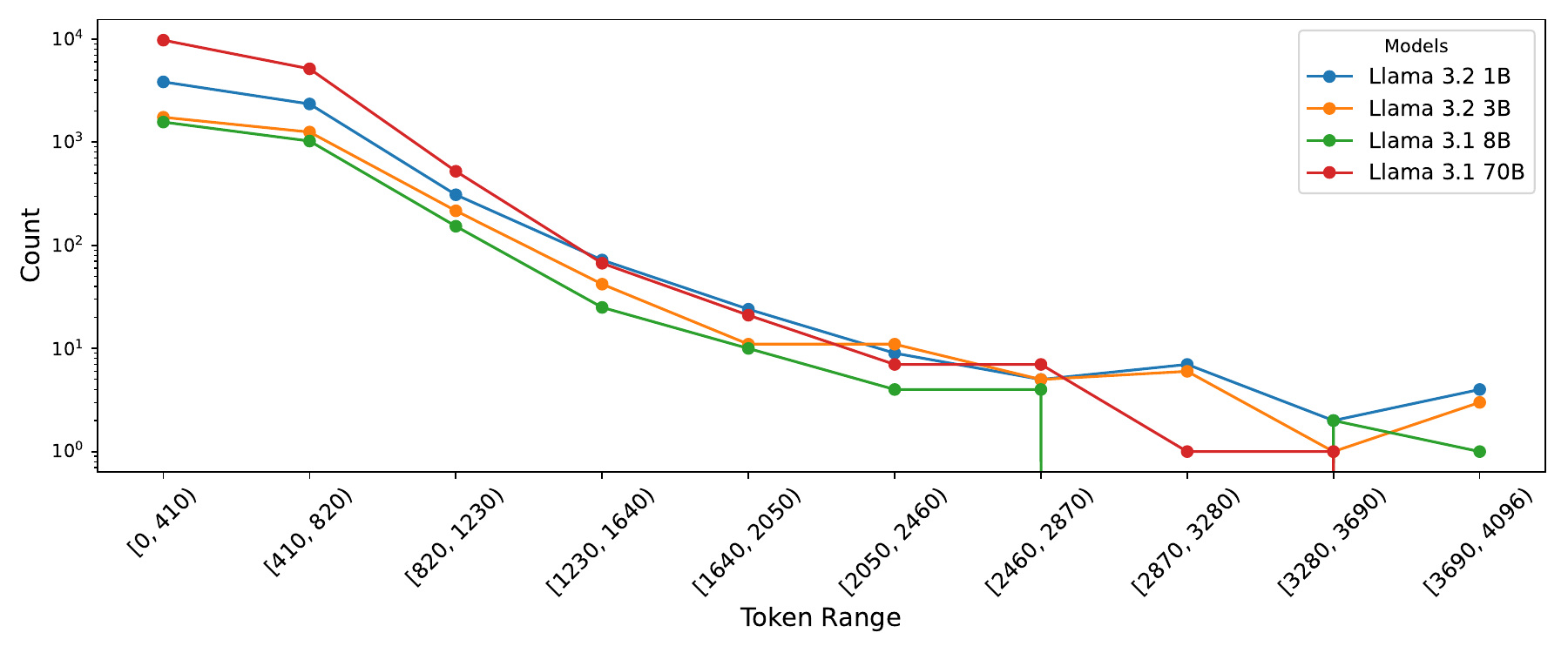}
    \caption{Per-model output token distribution for Chatbot Arena Dataset.}
    \label{fig:per_model_output_tokens}
\end{figure}

\subsubsection{Correlation Metrics for Generative AI}

We focus on the inference phase for Generative AI, which is typically the most computationally demanding part of a real-time user-interactive workload. Table~\ref{tab:spearman_correlations_genai} illustrates the Spearman correlations between total energy consumption and several key metrics for Large Language Model (LLM) inference experiments on Hardware Configuration~4 (HC-4). Separate GPU-only correlations are omitted here, having been verified to align closely with total energy usage (i.e., no additional insights are gleaned by isolating the GPU alone). 

\medskip
\noindent
We define seven core metrics that characterise model complexity and operational efficiency in LLM settings:
\begin{enumerate}
    \item \textbf{\texttt{energy\_per\_sample}}: 
    Represents the total average energy consumed for one LLM inference request. Since this serves as our baseline measure of energy usage, its correlation with total energy is, by definition, equal to 1.00.

    \item \textbf{\texttt{flops}}: 
    The total number of floating-point operations required for the model’s forward pass. This metric reflects the global computational cost of generating an inference output.

    \item \textbf{\texttt{model\_size\_to\_ram}}: 
    Compares the on-GPU size of the model to the total VRAM available, impacting caching efficiency and concurrency.

    \item \textbf{\texttt{parameters}}: 
    The full parameter count for the LLM reflects the overall model scale. 
    Larger models tend to require more computing but can be more expressive.

    \item \textbf{\texttt{request\_rate}}: 
    The number of inference RPS.  Higher RPS often leads to improved batching on GPUs, thus reducing per-request energy overhead up to resource limits.

    \item \textbf{\texttt{cache\_hit\_rate}}: 
    Fraction of queries that leverage cached tokens (e.g., from matching prompt prefixes). Effective caching lowers redundant computation and helps reduce energy usage.

    \item \textbf{\texttt{average\_output\_token\_length}}: 
    Mean token length of the model’s generated responses. While it does increase inference steps, its effect on total energy is often secondary to batching or model-scale factors.
\end{enumerate}

\begin{table}[t]
    \centering
    \caption{Spearman Correlations of the energy per sample consumption and various metrics for Generative AI models.}
    \begin{tabular}{lc}
        \toprule
        \textbf{Metric}               & \textbf{HC-4} \\ 
        \midrule
        energy\_per\_sample           & 1.00      \\ 
        flops                         & 0.32      \\ 
        model\_size\_to\_ram          & 0.32      \\  
        parameters                    & 0.32      \\ 
        request\_rate                 & -0.95     \\
        average\_output\_token\_length & -0.26     \\
        cache\_hit\_rate              & -0.32     \\
        \bottomrule
    \end{tabular}
    \label{tab:spearman_correlations_genai}
\end{table}

From Table~\ref{tab:spearman_correlations_genai}, we see that \texttt{energy\_per\_sample} naturally attains a perfect correlation as it is the reference factor. Additionally, \texttt{flops}, \texttt{model\_size\_to\_ram}, and \texttt{parameters} exhibit identical moderate correlations (0.32), in part because of simplifications in the FLOPs/parameter estimation library used~\cite{calflops}. By contrast, \texttt{request\_rate} shows a strong negative correlation (\(-0.95\)), underlining the energy benefit of processing multiple requests concurrently via batching. A similarly negative correlation for \texttt{cache\_hit\_rate} (\(-0.32\)) indicates that leveraging pre-computed tokens reduces redundant operations and, thus, overall energy. Lastly, \texttt{average\_output\_token\_length} displays a weak negative correlation (\(-0.26\)), suggesting response length is a less critical driver of total energy use when compared to concurrency and caching dynamics. The negative correlation may seem counter-intuitive; However, this is a consequence of the training biases of the different Llama model sizes, which, with the chatbot arena dataset, the smaller models produced longer generations than the larger models, as can be observed in Fig.~\ref{fig:per_model_output_tokens} for Output Histogram.

\section{Discussion}\label{sec:discussion}
Starting with our initial observations for Discriminative AI (Sec.~\ref{subsec:init_statistics}), it is evident that each model's unique architecture limits the potential for cross-model generalisations. For instance, while one model's energy consumption may be low, there is no guarantee that another model with similar characteristics will exhibit comparable energy efficiency. Investigating specific architectural features and model layers could unveil patterns or principles influencing energy consumption, paving the way for broader insights. However, when orchestrating a model deployment, it was evident (Fig.~\ref{fig:gpu_ram}) that a placement leading to the hardware being close to its saturation point (but not exceeding that) can lead to the best energy-performance result. This observation is shared across both Discriminative and Generative AI experiments.

As illustrated in Fig.~\ref{fig:loss_power}, energy reduction often outweighs accuracy gains in practical scenarios. Interestingly, training and inference durations are not directly correlated, rendering cross-phase or cross-hardware energy estimations unreliable. Although a heuristic might suggest that training typically requires approximately three times the duration of inference for the same number of samples, this does not hold universally.

Since time and total energy consumption scale linearly, short-lived profiling (e.g., training for one epoch or inferring for a small number of samples) can be a reliable predictor of energy consumption for larger-scale scenarios. Moreover, models that achieve comparable accuracy but demonstrate faster runtimes can yield substantial long-term energy savings. Based on the energy split observed in Fig.~\ref{fig:combined_power} and taking into account Facebook's energy split presented in Sec.~\ref{subsec:energy_consumption}, prioritising models that are energy-efficient during inference is more beneficial for real-world applications than focusing solely on training energy efficiency. 

To refine energy consumption predictions, strategies that analyse initial learning curves in conjunction with power profiles can provide accurate estimates of total energy usage. Additionally, Fig.~\ref{fig:gpu_ram} demonstrates that hardware power profiles are not strictly linear. Manufacturers often push device limits for marginal performance gains, which can lead to inefficiencies. Techniques like power capping optimisation (e.g., \cite{frost}) can mitigate this issue and significantly reduce energy consumption.

Considering various computational efficiency metrics (Sec.~\ref{subsec:computation_metrics}), our findings, contrary to the literature, suggest that the ratio of MACs to model parameters (\textbf{\texttt{macs\_param}}) offers a more consistent and reliable predictor than the model's MACs. This is endorsed by the strong correlation observed across different hardware configurations (Table~\ref{tab:spearman_correlations}). Similarly, \textbf{\texttt{energy\_per\_sample}} emerges as a robust metric due to its direct temporal correlation with energy use and can be easily calculated with short-lived experiments. This is the case also for \textbf{\texttt{overall\_efficiency}}  -- defined as the ratio of accuracy, throughput, and system utilisation -- that again can be used for long-term estimations, particularly for cases where ML models force the hardware to operate close to its saturation point. 

Finally, all the above metrics assume access to the energy consumption of the hardware. When such measurements are not available, predictive models could be built based on computational efficiency metrics, model hyperparameters, and hardware characteristics, which could effectively estimate the expected energy consumption. Excluding all energy-related metrics, we ran a Lasso regression to select the most important features for that. Our dataset was created by combining the measurements across all hardware configurations and models, and our train-test split was $80\%:20\%$. From this investigation, the most important features chosen were the GPU's memory utilisation, the MACs per parameter, the \textbf{\texttt{work\_done}}, the \textbf{\texttt{model\_size\_to\_ram}}, the MACs and the model size, with a combined importance of $\approx 65\%$. To that extent, a large investigation of multiple hardware configurations and models can create a very interesting dataset for the community that can be leveraged for future energy-efficient ML investigations. Moving on to the Generative AI experiments, we conducted a similar Lasso regression investigation. For this investigation, we also considered the cache hit rate to take into account the cached tokens and what might happen in higher load scenarios. The most influential and negative factor is RPS, confirming that batching/multithreading is key to energy efficiency, and overall, the RPS, the cache hit rate, and the average output tokens with combined importance of  $\approx 75\%$.

Our Generative AI findings suggest that, although larger models (e.g., 70B) provide improved capabilities, they also incur significantly higher energy costs per request, especially at lower RPS rates where resource utilisation is less efficient (Fig.~\ref{fig:total_energy_per_request}). The Energy Per Output token for the different models shows a similar trend in Fig.~\ref{fig:energy_per_output_token}. Furthermore, in Fig.~\ref{fig:per_device_energy}, we saw that the CPU consumption of different model sizes per request completed does not vary wildly between model sizer for a given hardware and a given RPS rate, whilst the GPU consumption does vary significantly. For sustainable deployments, this indicates that choosing appropriately sized models based on the anticipated RPS and computational requirements can lead to substantial energy savings. For applications with predictable and moderate request rates, smaller models in the range of 1-3 billion parameters offer an advantageous balance between performance and energy efficiency. Furthermore, we can also observe that operating the servers closer to saturation capacity significantly decreases the energy cost per request due to the increased throughput in Tokens/second (as in the case of Discriminative AI). However, it is also important to note that the latency is also likely to increase the closer the server gets to saturation.

From a deployment perspective, larger models generally offer higher accuracy but at the cost of significantly greater energy and resource consumption. To address this, fine-tuning smaller models to achieve accuracy levels closer to those of larger models presents a viable approach to reducing these costs. This strategy not only enhances energy efficiency but also extends the long-term utility of the models.

Overall, and based on our findings, several practical implementation strategies and recommendations can be derived for industry practitioners aiming to deploy energy-efficient ML systems. For Discriminative models, selecting architectures such as ResNet or VGG - which show strong performance while consuming significantly less energy—can provide optimal trade-offs for real-time inference scenarios. Batch size tuning should be used judiciously, particularly in hardware-constrained environments, to avoid unnecessary power draw without compromising performance. For Generative models, our results show that smaller LLMs (e.g., 3B or 8B) can achieve high throughput and energy efficiency under moderate request loads, making them preferable for scalable inference workloads. Integrating energy profiling into MLOps or GenOps frameworks enables dynamic model selection, power capping, or adaptive inference based on operational requirements.

As our final thoughts, while our study provides a comprehensive empirical evaluation of energy consumption across various Discriminative and Generative AI models, we acknowledge that our investigation is based on a finite set of hardware configurations. Considering other architectures (e.g., edge devices or ARM-based architectures) and a larger set of hardware configurations will provide more comprehensive results and correlations on how different model architectures operate across different hardware configurations. Moreover, our Generative AI analysis concentrated solely on inference workloads using pre-trained models, excluding the training phase due to its substantial cost and limited accessibility for many practitioners. While we used real-world workloads and datasets, our study does not account for all possible application-specific optimisations, such as quantisation-aware training or adaptive model scaling at runtime. Finally, our study focused on the model parameters but not so much on the individual layers of each model. An investigation targeting the energy consumption of different model layer types (e.g., convolutional, activation, pooling, etc.) will give more practical guidelines to ML practitioners who aim to build energy-efficient models. All the above limitations could be addressed in future research activities. Finally, integrating all the above practices in a real-world MLOps or GenOps pipeline will reveal more areas of consideration that can enhance the energy efficiency of such a system and enable more practical real-world impact and adoption by industry practitioners.

\section{Conclusions}\label{sec:conclusion}
This study underscores the importance of energy-efficient practices in both Discriminative and Generative AI models, providing empirical insights that challenge common assumptions about energy consumption patterns. For Discriminative models, we show that optimising model architecture, hyperparameters, and hardware provisioning can yield significant energy savings without compromising performance, often surpassing the benefits of marginal accuracy improvements. In Generative AI, particularly with LLMs, balancing model size and reasoning with request-handling capability emerges as a crucial factor for energy efficiency, where larger models may not increase energy demands as long as utilisation is low. Our findings highlight that energy consumption dynamics vary significantly across training, inference, and hardware configurations, emphasising the necessity for tailored strategies within each ML pipeline stage. Ultimately, this study demonstrates that with informed choices around model design, configuration, and deployment, AI/ML systems can be developed in alignment with environmental sustainability. By establishing a robust framework for energy-conscious ML operations, this work lays the groundwork for future research and industry practices to minimise the environmental impact of AI advancements. However, our study is limited to a select number of models and hardware platforms and does not cover edge devices or pipeline-level dynamic optimisations. Future work could explore adaptive strategies for energy management, real-time deployment considerations, and broader hardware-software co-design approaches to further improve sustainability in ML pipelines.

\authorcontributions{Conceptualization, A.S.-M., I. M., P. L., K. K, A. K.; methodology, A.S.-M., P. L., I. M., K. K, A. K.; software, A.S.-M., I. M.; validation, A.S.-M., I. M.; formal analysis, A.S.-M., P. L., I. M., A. K.; investigation, A.S.-M., P. L., I. M., A. K.; resources, A.K.; data curation, A.S.-M., I. M.; writing---original draft preparation, A.S.-M., P. L., I. M., A. K.; writing---review and editing, I. M., P. L. K. K. A. K.; visualization, A.S.-M., I. M.; supervision, A. K.; project administration, A. K.; funding acquisition, I. M, K.K., A. K. All authors have read and agreed to the published version of the manuscript.}

\funding{This work was funded in part by Toshiba Europe Ltd. and Bristol Research and Innovation Laboratory (BRIL). This work is also a contribution by Project REASON, a UK Government funded project under the Future Open Networks Research Challenge (FONRC) sponsored by the Department of Science Innovation and Technology (DSIT).}

\institutionalreview{Not applicable.}

\informedconsent{Not applicable.}

\dataavailability{The datasets presented in this article are readily available from the community. The software built is not readily available because of Toshiba's internal data/software control policies. Requests to access the software should be directed to Aftab Khan (aftab.khan@toshiba-bril.com)}

\conflictsofinterest{Authors are employed by Toshiba Europe Ltd./Digital Catapult. The remaining authors declare that the research was conducted in the absence of any commercial or financial relationships that could be construed as a potential conflict of interest.} 

\begin{adjustwidth}{-\extralength}{0cm}

\reftitle{References}



\begin{thebibliography}{999}



\bibitem[Luccioni et~al.(2020)Luccioni, Lacoste, and Schmidt]{co2Emissions}
Luccioni, A.; Lacoste, A.; Schmidt, V.
\newblock {Estimating Carbon Emissions of Artificial Intelligence [Opinion]}.
\newblock {\em IEEE Technol. Soc. Mag} {\bf 2020}, {\em 39},~48--51.

\bibitem[Kathikeyan et~al.(2022)Kathikeyan, Revathi, Supreeth, Sasidevi, Ahmed,
  and Das]{immersiveMedia}
Kathikeyan, T.; Revathi, S.; Supreeth, B.R.; Sasidevi, J.; Ahmed, M.; Das, S.
\newblock {Artificial Intelligence and Mixed Reality Technology for Interactive
  Display of Images in Smart Area}.
\newblock In Proceedings of the 2022 5th International Conference on Contemporary Computing and Informatics (IC3I), Uttar Pradesh, India, 14--16 December 2022; pp. 2049--2053.
\newblock {\url{https://doi.org/10.1109/IC3I56241.2022.10072411}}.

\bibitem[Moinnereau et~al.(2022)Moinnereau, de~Oliveira, and
  Falk]{Moinnereau2022}
Moinnereau, M.A.; de~Oliveira, A.A.; Falk, T.H.
\newblock {Immersive Media Experience: A Survey of Existing Methods and Tools
  for Human Influential Factors Assessment}.
\newblock {\em Qual. User Exp.} {\bf 2022}, {\em 7},~5.

\bibitem[Bertolini et~al.(2021)Bertolini, Mezzogori, Neroni, and
  Zammori]{bertolini2021machine}
Bertolini, M.; Mezzogori, D.; Neroni, M.; Zammori, F.
\newblock Machine Learning for industrial applications: A comprehensive
  literature review.
\newblock {\em Expert Syst. Appl.} {\bf 2021}, {\em 175},~114820.

\bibitem[Wang et~al.(2023)Wang, Pan, Yan, Su, and Luan]{genAI}
Wang, Y.; Pan, Y.; Yan, M.; Su, Z.; Luan, T.H.
\newblock {A Survey on ChatGPT: AI–Generated Contents, Challenges, and
  Solutions}.
\newblock {\em IEEE Open J. Comput. Soc} {\bf 2023}, {\em 4},~280--302.

\bibitem[Li et~al.(2024)Li, S{\'a}nchez-Momp{\'o}, Farnham, Khan, and
  Aijaz]{li2024large}
Li, P.; S{\'a}nchez-Momp{\'o}, A.; Farnham, T.; Khan, A.; Aijaz, A.
\newblock Large Generative AI Models meet Open Networks for 6G: Integration,
  Platform, and Monetization.
\newblock {\em arXiv} {\bf 2024}, arXiv:2410.18790.

\bibitem[Katsaros et~al.(2024)Katsaros, Mavromatis, Antonakoglou, Ghosh,
  Kaleshi, Mahmoodi, Asgari, Karousos, Tavakkolnia, Safi, Hass, Vrontos, Emami,
  Ullauri, Moazzeni, and Simeonidou]{aiNative6G}
Katsaros, K.; Mavromatis, I.; Antonakoglou, K.; Ghosh, S.; Kaleshi, D.;
  Mahmoodi, T.; Asgari, H.; Karousos, A.; Tavakkolnia, I.; Safi, H.;  et~al.
\newblock {AI-Native Multi-Access Future Networks---The REASON Architecture}.
\newblock {\em IEEE Access} {\bf 2024}, \emph{12}, 178586--178622.

\bibitem[Patterson et~al.(2022)Patterson, Gonzalez, Hölzle, Le, Liang,
  Munguia, Rothchild, So, Texier, and Dean]{mlStrategies}
Patterson, D.; Gonzalez, J.; Hölzle, U.; Le, Q.; Liang, C.; Munguia, L.M.;
  Rothchild, D.; So, D.R.; Texier, M.; Dean, J.
\newblock {The Carbon Footprint of Machine Learning Training Will Plateau, Then
  Shrink}.
\newblock {\em Computer} {\bf 2022}, {\em 55},~18--28.

\bibitem[Schwartz et~al.(2020)Schwartz, Dodge, Smith, and Etzioni]{greenAI}
Schwartz, R.; Dodge, J.; Smith, N.A.; Etzioni, O.
\newblock {Green AI}.
\newblock {\em Commun. ACM} {\bf 2020}, {\em 63},~54–63.

\bibitem[Verdecchia et~al.(2023)Verdecchia, Sallou, and Cruz]{systematicReview}
Verdecchia, R.; Sallou, J.; Cruz, L.
\newblock {A Systematic Review of Green AI}.
\newblock {\em WIREs Data Min. Knowl. Discov.} {\bf 2023}, {\em
  13},~e1507.

\bibitem[Singh et~al.(2024)Singh, Patel, Ehtesham, Kumar, and
  Khoei]{llmSustainabilitySurvey}
Singh, A.; Patel, N.P.; Ehtesham, A.; Kumar, S.; Khoei, T.T.
\newblock {A Survey of Sustainability in Large Language Models: Applications,
  Economics, and Challenges}.
\newblock {\em arXiv} {\bf 2024}, arXiv:2412.04782.

\bibitem[Yang et~al.(2017)Yang, Chen, and Sze]{energyAwarePruning}
Yang, T.J.; Chen, Y.H.; Sze, V.
\newblock {Designing Energy-Efficient Convolutional Neural Networks Using
  Energy-Aware Pruning}. In Proceedings of the 2017 IEEE Conference on Computer Vision and Pattern Recognition (CVPR), Honolulu, HI, USA, \mbox{21--26~July~2017;} pp. 6071--6079.

\bibitem[Eliezer et~al.(2023)Eliezer, Banner, Ben-Yaakov, Hoffer, and
  Michaeli]{energyAwareQuantisation}
Eliezer, N.S.; Banner, R.; Ben-Yaakov, H.; Hoffer, E.; Michaeli, T.
\newblock {Power Awareness In Low Precision Neural Networks}.
\newblock In Proceedings of the Computer Vision---ECCV 2022 Workshops, Tel Aviv, Israel, 23--27 October  2022; pp. 67--83.

\bibitem[de~Reus et~al.(2024)de~Reus, Oprescu, and Zuidema]{llmPruning}
de~Reus, P.; Oprescu, A.; Zuidema, J.
\newblock {An Exploration of the Effect of Quantisation on Energy Consumption
  and Inference Time of StarCoder2}.
\newblock {\em arXiv} {\bf 2024}, {\em arXiv:2411.12758}.

\bibitem[Cottier et~al.(2024)Cottier, Rahman, Fattorini, Maslej, and
  Owen]{cottier2024risingcoststrainingfrontier}
Cottier, B.; Rahman, R.; Fattorini, L.; Maslej, N.; Owen, D.
\newblock The rising costs of training frontier AI models.
\newblock {\em arXiv} {\bf 2024}, arXiv:2405.21015.

\bibitem[Touvron et~al.(2023)Touvron, Lavril, Izacard, Martinet, Lachaux,
  Lacroix, Rozière, Goyal, Hambro, Azhar, Rodriguez, Joulin, Grave, and
  Lample]{touvron2023llamaopenefficientfoundation}
Touvron, H.; Lavril, T.; Izacard, G.; Martinet, X.; Lachaux, M.A.; Lacroix, T.;
  Rozière, B.; Goyal, N.; Hambro, E.; Azhar, F.;  et~al.
\newblock LLaMA: Open and Efficient Foundation Language Models.
\newblock {\em arXiv} {\bf 2023}, arXiv:2302.13971.


\bibitem[Wu et~al.(2022)Wu, Raghavendra, Gupta, Acun, Ardalani, Maeng, Chang,
  Aga, Huang, Bai, Gschwind, Gupta, Ott, Melnikov, Candido, Brooks, Chauhan,
  Lee, Lee, Akyildiz, Balandat, Spisak, Jain, Rabbat, and
  Hazelwood]{facebookAI}
Wu, C.J.; Raghavendra, R.; Gupta, U.; Acun, B.; Ardalani, N.; Maeng, K.; Chang,
  G.; Aga, F.; Huang, J.; Bai, C.;  et~al.
\newblock {Sustainable AI: Environmental Implications, Challenges and
  Opportunities}. \emph{Proc. Mach. Learn. Syst.} \textbf{2022}, \emph{4}, 795--813.

\bibitem[Islam et~al.(2023)Islam, Zisad, Kor, and Hasan]{mlModelSustainability}
Islam, M.S.; Zisad, S.N.; Kor, A.L.; Hasan, M.H.
\newblock {Sustainability of Machine Learning Models: An Energy Consumption
  Centric Evaluation}.
\newblock In Proceedings of the 2023 International Conference on Electrical, Computer and Communication Engineering (ECCE), Chittagong, Bangladesh, 23--25 February 2023; pp. 1--6.

\bibitem[Strubell et~al.(2020)Strubell, Ganesh, and
  McCallum]{Strubell_Ganesh_McCallum_2020}
Strubell, E.; Ganesh, A.; McCallum, A.
\newblock {Energy and Policy Considerations for Modern Deep Learning Research}. \emph{Proc. AAAI Conf. Artif. Intell.} \textbf{2020}, \emph{34}, 13693--13696.

\bibitem[Samsi et~al.(2023)Samsi, Zhao, McDonald, Li, Michaleas, Jones,
  Bergeron, Kepner, Tiwari, and Gadepally]{llmLlamaPowerCaping}
Samsi, S.; Zhao, D.; McDonald, J.; Li, B.; Michaleas, A.; Jones, M.; Bergeron,
  W.; Kepner, J.; Tiwari, D.; Gadepally, V.
\newblock {From Words to Watts: Benchmarking the Energy Costs of Large Language
  Model Inference}.  In Proceedings of the 2023 IEEE High Performance Extreme Computing Conference (HPEC), Boston, MA, USA, 25--29 September 2023; pp. 1--9.
\newblock {\url{https://doi.org/10.1109/HPEC58863.2023.10363447}}.

\bibitem[Husom et~al.(2024)Husom, Goknil, Shar, and Sen]{llmEnergy}
Husom, E.J.; Goknil, A.; Shar, L.K.; Sen, S.
\newblock {The Price of Prompting: Profiling Energy Use in Large Language
  Models Inference}.
\newblock {\em arXiv} {\bf 2024},  arXiv:2410.18790.

\bibitem[Li et~al.(2024)Li, Mavromatis, Farnham, Aijaz, and
  Khan]{li2024adapting}
Li, P.; Mavromatis, I.; Farnham, T.; Aijaz, A.; Khan, A. Adapting MLOps for Diverse In-Network Intelligence in 6G Era: Challenges and Solutions. {\em arXiv} {\bf 2024}, arXiv:2410.18793.

\bibitem[Testi et~al.(2022)Testi, Ballabio, Frontoni, Iannello, Moccia, Soda,
  and Vessio]{mlOpsTaxonomy}
Testi, M.; Ballabio, M.; Frontoni, E.; Iannello, G.; Moccia, S.; Soda, P.;
  Vessio, G.
\newblock {MLOps: A Taxonomy and a Methodology}.
\newblock {\em IEEE Access} {\bf 2022}, {\em 10},~63606--63618.

\bibitem[Teo et~al.(2024)Teo, Chua, Jasser, and Wong]{hybridML}
Teo, T.W.; Chua, H.N.; Jasser, M.B.; Wong, R.T.
\newblock {Integrating Large Language Models and Machine Learning for Fake News
  Detection}. In Proceedings of the 2024 20th IEEE International Colloquium on Signal Processing and Its Applications, CSPA 2024---Conference Proceedings, Langkawi, Malaysia, 1--2 March 2024;
pp. 102--107.

\bibitem[Satorras et~al.(2019)Satorras, Akata, and Welling]{hybridML2}
Satorras, V.G.; Akata, Z.; Welling, M.
\newblock Combining Generative and Discriminative Models for Hybrid Inference.
\newblock {\em arXiv} {\bf 2019}, arXiv:1906.02547.




\bibitem[Zhang et~al.(2024)Zhang, Du, Liu, Niyato, Kang, Xiong, Jamalipour, and
  In~Kim]{genAIAgents}
Zhang, R.; Du, H.; Liu, Y.; Niyato, D.; Kang, J.; Xiong, Z.; Jamalipour, A.;
  In~Kim, D.
\newblock {Generative AI Agents With Large Language Model for Satellite
  Networks via a Mixture of Experts Transmission}.
\newblock {\em IEEE J. Sel. Areas Commun.} {\bf 2024},
  {\em 42},~3581--3596. {\url{https://doi.org/10.1109/JSAC.2024.3459037}}.


\bibitem[{Mavromatis} et~al.(2024){Mavromatis}, {Katsaros}, and
  {Khan}]{energy_ml}
{Mavromatis}, I.; {Katsaros}, K.; {Khan}, A.
\newblock {Computing Within Limits: An Empirical Study of Energy Consumption in
  ML Training and Inference}.
\newblock In Proceedings of the International Scientific Conference on
Information, Communication and Energy Systems and Technologies (ICEST 2024)---Workshop on Artificial
Intelligence for Sustainable Development (ARISDE 2024), Sozopol, Bulgaria, 1--3 July 2024. 


\bibitem[Conti et~al.(2023)Conti, Jimenez, del Rio, Castano-Solis, Serrano, and
  Fraile-Ardanuy]{physicalMeter}
Conti, G.; Jimenez, D.; del Rio, A.; Castano-Solis, S.; Serrano, J.;
  Fraile-Ardanuy, J.
\newblock {A Multi-Port Hardware Energy Meter System for Data Centers and
  Server Farms Monitoring}.
\newblock {\em Sensors} {\bf 2023}, {\em 23}, 119.

\bibitem[Rinaldi et~al.(2019)Rinaldi, Bonafini, Ferrari, Flammini, Pasetti, and
  Sisinni]{hardwareSync}
Rinaldi, S.; Bonafini, F.; Ferrari, P.; Flammini, A.; Pasetti, M.; Sisinni, E.
\newblock {Software-based Time Synchronization for Integrating Power Hardware
  in the Loop Emulation in IEEE1588 Power Profile Testbed}.
\newblock In 2019 IEEE International Symposium on Precision Clock Synchronization for Measurement, Control, and Communication (ISPCS), Portland, OR, USA, 22--27 September 2019; \mbox{pp.~1--6.}

\bibitem[Lin et~al.(2021)Lin, Yu, Gao, Liu, Li, Fong, and Wang]{LIN20211045}
Lin, W.; Yu, T.; Gao, C.; Liu, F.; Li, T.; Fong, S.; Wang, Y.
\newblock {A Hardware-aware CPU Power Measurement Based on the Power-exponent
  Function model for Cloud Servers}.
\newblock {\em Inf. Sci.} {\bf 2021}, {\em 547},~1045--1065.

\bibitem[{NVIDIA Corporation}(2016)]{Nvidia2016}
{NVIDIA Corporation}.
\newblock nvidia-smi.txt,  2016. 


\bibitem[{Katsenou} et~al.(2022){Katsenou}, {Mao}, and
  {Mavromatis}]{erqTradeOffVideoCodecs}
{Katsenou}, A.; {Mao}, J.; {Mavromatis}, I.
\newblock {Energy-Rate-Quality Tradeoffs of State-of-the-Art Video Codecs}.
\newblock In Proceedings of the 2022~Picture Coding Symposium (PCS), San Jose, CA, USA, 7--9 December 2022; pp. 265--269.

\bibitem[Vogelsang(2010)]{dramPowerConsumption}
Vogelsang, T.
\newblock {Understanding the Energy Consumption of Dynamic Random Access
  Memories}. In Proceedings of the Annual IEEE/ACM International Symposium on Microarchitecture, Atlanta, GA, USA, 4--8 December 2010; pp. 363--374.

\bibitem[Teo and Chia(2018)]{imageClassification}
Teo, J.; Chia, J.T.
\newblock {Deep Neural Classifiers For Eeg-Based Emotion Recognition In
  Immersive Environments}.
\newblock In Proceedings of the 2018 International Conference on Smart Computing and Electronic Enterprise (ICSCEE), Shah Alam, Malaysia, 11--12 July 2018; pp. 1--6.

\bibitem[Gaona-Garcia et~al.(2019)Gaona-Garcia, Montenegro-Marin, de~Inigo
  Sarría Martínez~Mendivil, Rodríguez, and Riano]{educationalGames}
Gaona-Garcia, P.A.; Montenegro-Marin, C.E.; de~Inigo Sarría
  Martínez~Mendivil.; Rodríguez, A.O.R.; Riano, M.A.
\newblock {Image Classification Methods Applied in Immersive Environments for
  Fine Motor Skills Training in Early Education}.
\newblock {\em Int. J. Interact. Multi.} {\bf 2019}, {\em 5},~151--158.

\bibitem[Krizhevsky(2009)]{cifar10}
Krizhevsky, A. \emph{Learning Multiple Layers of Features from Tiny Images}; University of Toronto: Toronto, ON, Canada, 2009. 

\bibitem[{Mavromatis} et~al.(2023){Mavromatis}, {De Feo}, {Carnelli},
  {Piechocki}, and {Khan}]{frost}
{Mavromatis}, I.; {De Feo}, S.; {Carnelli}, P.; {Piechocki}, R.J.; {Khan}, A.
\newblock {FROST: Towards Energy-efficient AI-on-5G Platforms---A GPU Power
  Capping Evaluation}.
\newblock In Proceedings of the 2023 IEEE Conference on Standards for Communications and Networking (CSCN), Munich, Germany, 6--8 November 2023; pp.~1--6.

\bibitem[Aldin and Aldin(2022)]{batchSizeComparison}
Aldin, N.B.; Aldin, S.S.A.B.
\newblock {Accuracy Comparison of Different Batch Size for a Supervised Machine
  Learning Task with Image Classification}.
\newblock In Proceedings of the 2022 9th International Conference on Electrical and Electronics Engineering (ICEEE), Alanya, Turkey, 29--31 March 2022; pp.~316--319.

\bibitem[Kwon et~al.(2023)Kwon, Li, Zhuang, Sheng, Zheng, Yu, Gonzalez, Zhang,
  and Stoica]{kwon2023efficient}
Kwon, W.; Li, Z.; Zhuang, S.; Sheng, Y.; Zheng, L.; Yu, C.H.; Gonzalez, J.E.;
  Zhang, H.; Stoica, I.
\newblock {Efficient Memory Management for Large Language Model Serving with
  PagedAttention}. In Proceedings of the 29th Symposium on Operating Systems Principles, Koblenz, Germany, 23--26 October 2023.

\bibitem[Zheng et~al.(2023)Zheng, Chiang, Sheng, Zhuang, Wu, Zhuang, Lin, Li,
  Li, Xing, Zhang, Gonzalez, and Stoica]{zheng2023judging}
Zheng, L.; Chiang, W.L.; Sheng, Y.; Zhuang, S.; Wu, Z.; Zhuang, Y.; Lin, Z.;
  Li, Z.; Li, D.; Xing, E.P.;  et~al.
\newblock {Judging LLM-as-a-judge with MT-Bench and Chatbot Arena}.
\newblock {\em arXiv} {\bf 2023}, arXiv:2306.05685.

\bibitem[xiaoju ye(2023)]{calflops}
Ye, X. calflops: A FLOPs and Params calculate tool for neural networks in
  pytorch framework, 2023.


\end{thebibliography}
\PublishersNote{}
\end{adjustwidth}
\end{document}